\documentclass[10pt,twocolumn,letterpaper]{article}

\usepackage{cvpr}              % To produce the CAMERA-READY version
\usepackage[dvipsnames]{xcolor}

\definecolor{cvprblue}{rgb}{0.21,0.49,0.74}
\usepackage[pagebackref,breaklinks,colorlinks,citecolor=cvprblue]{hyperref}
\usepackage{epsfig}
\usepackage{amssymb}
\usepackage{times} 
\usepackage{url}
\usepackage{graphicx}
\usepackage{algorithm}
\usepackage{algorithmic}
\usepackage{amsmath}
\usepackage{multirow}
\usepackage{makecell}
\usepackage{setspace}
\usepackage{booktabs}
\usepackage{subfloat}
\usepackage{subcaption}
\usepackage{lipsum}
\usepackage{xcolor}

\newcommand{\bsk}[1]{\textcolor{black}{#1}}
\newcommand{\sk}[1]{\textcolor{black}{#1}}

\def\paperID{13655} % *** Enter the Paper ID here
\def\confName{CVPR}
\def\confYear{2024}

\title{DiPrompT: Disentangled Prompt Tuning for \\ Multiple Latent Domain Generalization in Federated Learning}

\author{Sikai Bai\textsuperscript{\rm 1$\ast$}, Jie Zhang\textsuperscript{\rm 2$\ast$}, Shuaicheng Li\textsuperscript{\rm 3 $\dag\ddag$}, \\ Song Guo\textsuperscript{\rm 1$\dag$},  Jingcai Guo\textsuperscript{\rm 2}, Jun Hou\textsuperscript{\rm 3}, Tao Han\textsuperscript{\rm 1}, Xiaocheng Lu\textsuperscript{\rm 1} \\
\textsuperscript{\rm 1}The Hong Kong University of Science and Technology,
\textsuperscript{\rm 2}The Hong Kong Polytechnic University, \\
\textsuperscript{\rm 3}SenseTime Research
}

\begin{document}

\maketitle
\begin{abstract}
Federated learning (FL) has emerged as a powerful paradigm for learning from decentralized data, and federated domain generalization further considers the test dataset (target domain) is absent from the decentralized training data (source domains).  
However, most existing FL methods assume that domain labels are provided during training, and their evaluation imposes explicit constraints on the number of domains, which must strictly match the number of clients. Because of the underutilization of numerous edge devices and additional cross-client domain annotations in the real world, such restrictions may be impractical and involve potential privacy leaks.
In this paper, we propose an efficient and novel approach, called \textbf{Di}sentangled \textbf{Promp}t \textbf{T}uning (\textbf{DiPrompT}),
a method that tackles the above restrictions by learning adaptive prompts for domain generalization in a distributed manner.
%, based on prompt tuning. It tackles the above restrictions by extracting diversified complementary knowledge and obtains better target domain prediction in decentralized distribution.}
Specifically, we first design two types of prompts,  i.e., global prompt to capture general knowledge across all clients and domain prompts to capture domain-specific knowledge. They eliminate the restriction on the one-to-one mapping between source domains and local clients.
\bsk{Furthermore, a dynamic query metric is introduced to automatically search the suitable domain label for each sample,  which includes two-substep text-image alignments based on prompt tuning without labor-intensive annotation.}
Extensive experiments on multiple datasets demonstrate that our DiPrompT achieves superior domain generalization performance over state-of-the-art FL methods when domain labels are not provided, and even outperforms many centralized learning methods using domain labels.
\end{abstract}

%In this paper, we propose an efficient and novel approach, called Disentagled Prompts (DiPrompT), based on prompt learning to extract diversified source knowledge from complementary prompts for better target domain prediction in decentralized distribution, regardless of above restrictions. 

% \bsk{Specifically, we first design two types of prompts, i.e., global prompt and domain prompts, to eliminate the restriction on the one-to-one mapping between source domains and local clients. Global prompt maintains general knowledge across all clients, while domain prompts separately learn each domain-specific knowledge from multiple clients and explore the relationship between different domains.}

%Specifically, we first design two types of prompts, i.e., global prompt and domain prompt, to maintain general knowledge across all clients, \bsk{while separately learning each domain-specific knowledge from multiple clients and exploring the relationship between different domains.}
%explore the relationship between different domains by learning domain-specific knowledge in multiple clients. 
%DiPrompT eliminates the restriction on the one-to-one mapping between source domains and local clients. 
\let\thefootnote\relax\footnotetext{$\ast$ Equal contribution, $\dag$ Corresponding authors, $\ddag$ Project leader}
\section{Introduction}
\label{sec:intro}
Federated learning (FL) is an emerging privacy-preserving machine-learning technique \citep{mcmahan2017communication}, which enables multiple clients (e.g., mobile devices or organizations) to collaboratively learn a global model without exchanging their private data. 
%
%However, a practical concern with conventional FL methods is that they only considered category heterogeneity in the data distributed across clients, and usually ignored the possible domain shift between training data (source domains) and test data (target domain) \cite{bai2023benchmarking}, which can incur poor performance on unseen target domains due to domain discrepancies.
However, a practical concern with conventional FL methods is that they usually ignore the possible domain shift between training data (source domains) and test data (target domain) \cite{bai2023benchmarking}, which can incur poor performance on unseen target domains due to domain discrepancies.

\begin{figure}[t]
   \centering
   \includegraphics[width=0.95\linewidth]{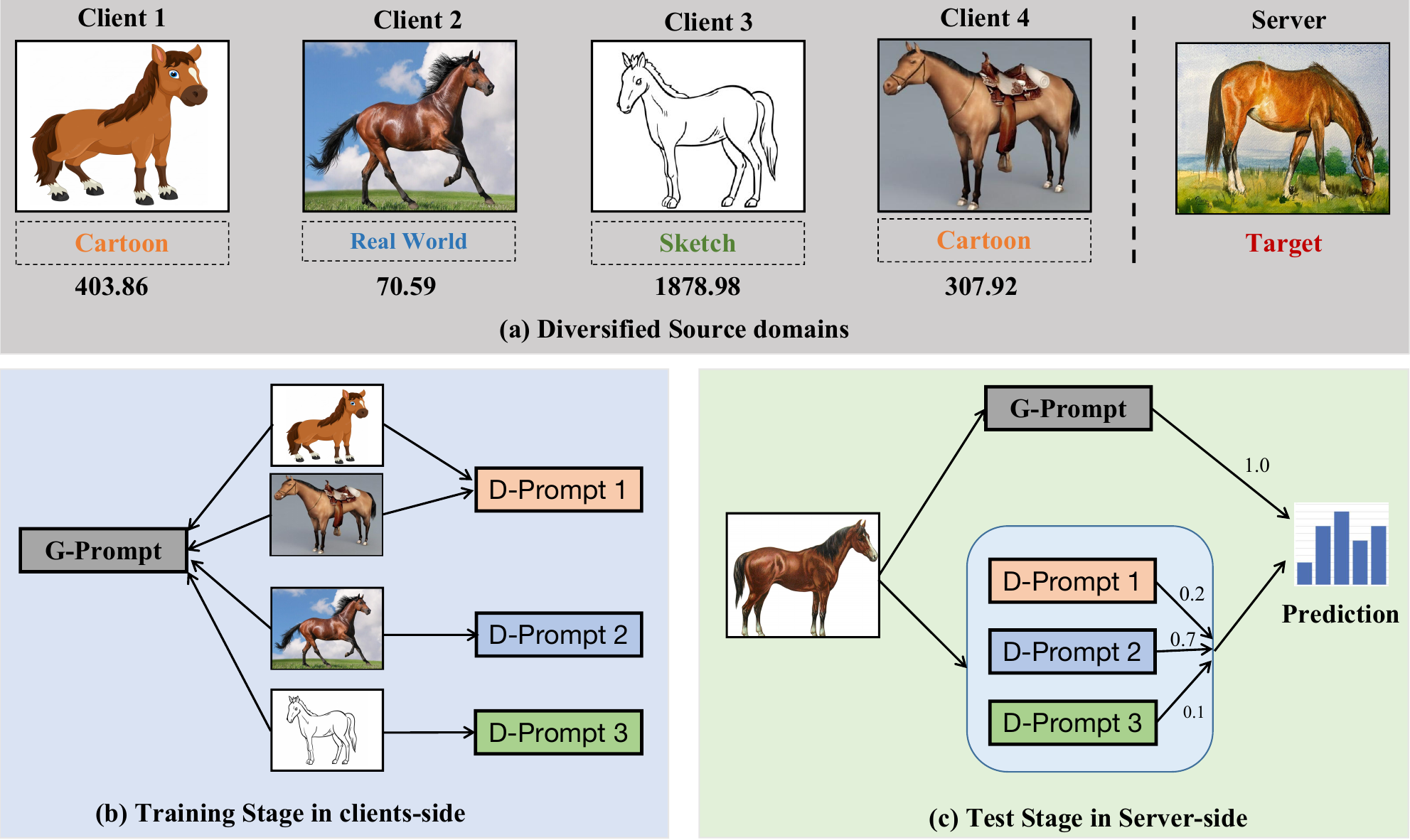}
   \caption[Caption for LOF]{The motivation example and our main idea. (a) When clients outnumber source domains, learning domain-invariant features may become challenging due to imbalanced contributions across domains/clients. Note that the contribution imbalance of local data is measured through its feature distances with the target domain.
   %The motivation example and our main idea. (a) When clients outnumber source domains,  learning domain-invariant features may become challenging. The imbalanced contribution of different clients among different domains also exacerbates the learning difficulty. Note that the contribution imbalance of local data is measured through its feature distances with the target domain.
   %When the number of clients exceeds the number of source domains, learning domain-invariant features may become challenging, and the imbalanced contribution of different clients among different domains also exacerbates the learning difficulty. Note that we measure contribution imbalance via feature distances with the target domain.
   %some clients that has small feature distance can provide more valuable knowledge for predicting target prediction. 
   (b) DiPrompT separates domain-specific features and general knowledge during local training. (c) DiPrompT adaptive ensembles for generic and valuable specific knowledge for better target domain prediction during inference.}
   \label{fig1}
\end{figure}
%To mitigate the above train-test domain shift, 
%domain generalization has been extensively studied in the centralized setting, which aims to train a generalized model across multiple source domains that perform well for the unseen target domain \cite{zhou2022domain}. 
%However, it is not straightforward to incorporate existing DG techniques into this FL setting, since most centralized methods require sharing the data representation across domains and they ignore the privacy constraints in FL scenarios. 

Recently, some research efforts have attempted to incorporate domain generalization into the FL framework. For example, FedDG \cite{liu2021feddg} shares the amplitude spectrum of images among local clients for medical image segmentation.  FedSR \cite{nguyen2022fedsr} builds a simple algorithm that utilizes two local regularizers for domain generalization. These methods extract domain-invariant features across all source domains.
Nevertheless, most of these methods only focus on domain-invariant features across clients, and they rely on the assumption of one-to-one mapping of client and domain.
%most of these methods focus solely on domain-invariant features across clients, and they 
%require that the number of clients must be equal to the number of domains. 
%If the number of local clients increases regardless of source domains, we find a new, more realistic challenge where data from a source domain is distributed across multiple clients, and some general knowledge may become specific due to only appearing in certain clients.
\sk{When the quantity of local clients increases regardless of source domains,  a more severe and real-world challenge emerges. In this scenario, data from a source domain can be dispersed among multiple clients, resulting in initially general knowledge that may be limited to certain clients and become specific.}
Thus learning sufficient invariant knowledge across all clients becomes infeasible. In Figure \ref{fig1}(a), for the horse class, the only general thing across all clients is shape. However, \bsk{some specific knowledge from certain clients possess a smaller feature distance with the target domain than others (e.g., the horse features distance between client 2 and the target domain is only 70.59).} Intuitively, we can additionally harness these features throughout the training phase to facilitate performance in the target domain.
%Such that they can aid object recognition in the target domain, and we want to retain them during training. 
Unfortunately, due to interference from spurious specific information and the entanglement between generic and specific features, it is non-trivial to separately extract generic and specific knowledge for each domain with a single model.
%and some source domains may contain more valuable information for the target domain
%most of them only focus on domain-agnostic feature representation across all clients, but when local clients become more diverse, learning domain-invariant knowledge becomes too casual and difficult. An example is shown in Figure \ref{fig1}(a), where the only general thing across all clients for the horse class seems to be shape. However, some domain-specific knowledge across a part of clients is also useful for object recognition in the target painting domain (e.g. texture information in clients 2 and 4), which we want to maintain during the training. Unfortunately, due to the interference from many spurious domain-specific information and the entanglement between generic and specific features, it is non-trivial to extract generic as well as discriminative knowledge for each domain with a single model. 
Moreover, domain information is indispensable for each local sample in these methods, but it is prohibitively expensive and may risk privacy leakage to annotate to which domain each sample belongs in a decentralized setting.
%there are some impractical constraints in the current evaluation. For example, 
%and limits their applicable potential in many cross-device scenarios
%For instance, in many cross-device scenarios, the number of clients could be extremely large and even include 3.5 billion active Android phones.

% 2) Requirement for domain labels.
%3) Their generalization performance relies on model local computation and communication, which can produce unaffordable resource consumption under large foundation models.  
%Besides, for the first restriction, FedCLIP constructs an additional attention-based adapter in a large vision-language model to achieve fast generalization, but it still suffers from the latter limitations.

To tackle the above issues, we propose a novel and efficient method termed Disentangled Prompt Tuning (DiPrompT) for domain generalization in FL settings. 
%The motivation is to spin off domain-specific and general features from local training with different lightweight components. Ideally, we hope to avoid mutual interference between these features and filter irrelevant specific attributes. Thus, the key complementary information can be exploited by adaptive ensembling valuable specific and generic knowledge to better tackle the target domain.
The motivation is to simultaneously spin off domain-specific and general features with different lightweight components in local training, minimizing their interference and removing irrelevant specifics. Furthermore, we can gain crucial complementary information by adaptive ensembling of specific and generic knowledge for the target domain in the test stage.
%This allows adaptive ensembling of crucial complementary information for more effective handling of the target domain.
Specifically, DiPrompT first introduces two types of prompts:  1) \textbf{Global prompt (G-Prompt)} maintains the domain-invariant representation across all clients. %and ignores specific attributes from some source domains. 
It would be invariant to the domain shift brought by all clients in decentralized training. 2)\textbf{Domain prompts (D-Prompts)}: \sk{Inspired by prototype learning \cite{SnellSZ17}, we construct a prototypical prompt for each predefined source domain, encapsulating discriminative specific knowledge from each source domain into the respective prompts. It resolves the count consistency restriction between clients and domains by domain-wise optimization and aggregation, while enhancing cross-client representation for each domain.}
Furthermore, when domain labels are unknown during both training and test periods (i.e., latent domains),
%prior methods assume that the latent domain of images is reflected in their style or multiple latent domains can be divided via clustering algorithms. They usually assigned pseudo-domain labels for each sample, but cannot utilize valuable textual knowledge about latent domains. Instead, 
we design an adaptive query mechanism to explore the potential domain for each sample.  An additional prompt (\textbf{Q-prompt}) is introduced, which automatically queries the domain label for each sample from all possible 
%shares learnable vectors with all classes and underlying domains, while it is optimized by self-supervised regulation during training. Using Q-Prompt, 
options by image-text alignments after excluding the interference of semantic categories. Finally, at inference time, we leverage a collaborative ensemble metric to provide valuable complementary information from G-Prompt and D-Prompts for better target prediction. %Moreover, it is noted that small-scale operations of prompt learning can significantly reduce the computational and communication burden brought by large pre-trained models. 

%By separating domain-specific knowledge from general knowledge learning across all clients, and providing more valuable information for server-side target prediction. It Meanwhile, we build an adaptive query mechanism to automatically search the latent domain for each sample, which mitigates domain labels' dependence during domain generalization.
\begin{itemize}
    \item To the best of our knowledge, this is the first lightweight work that handles federated domain generalization through prompt tuning. 
    We aspire that our research and findings can offer a fresh perspective toward solving cutting-edge challenges in domain generalization and federated learning.
    \item We propose DiPrompT, a novel federated domain generalization framework based on prompt tuning. It removes two impractical restrictions and provides complementary knowledge for generalization on unseen target domains with small-scale operations.
    %(dependency on domain labels and one-to-one mapping between clients and domains.)
    %a novel federated domain generalization framework based on prompt learning. It
    \item Extensive experiments on multiple domain generalization tasks verify the superiority of DiPrompT over state-of-the-art methods, which even outperforms some domain generalization methods with domain labels.
\end{itemize}

\begin{figure*}
	\begin{center}
		\centering
		\includegraphics[width=0.75\linewidth]{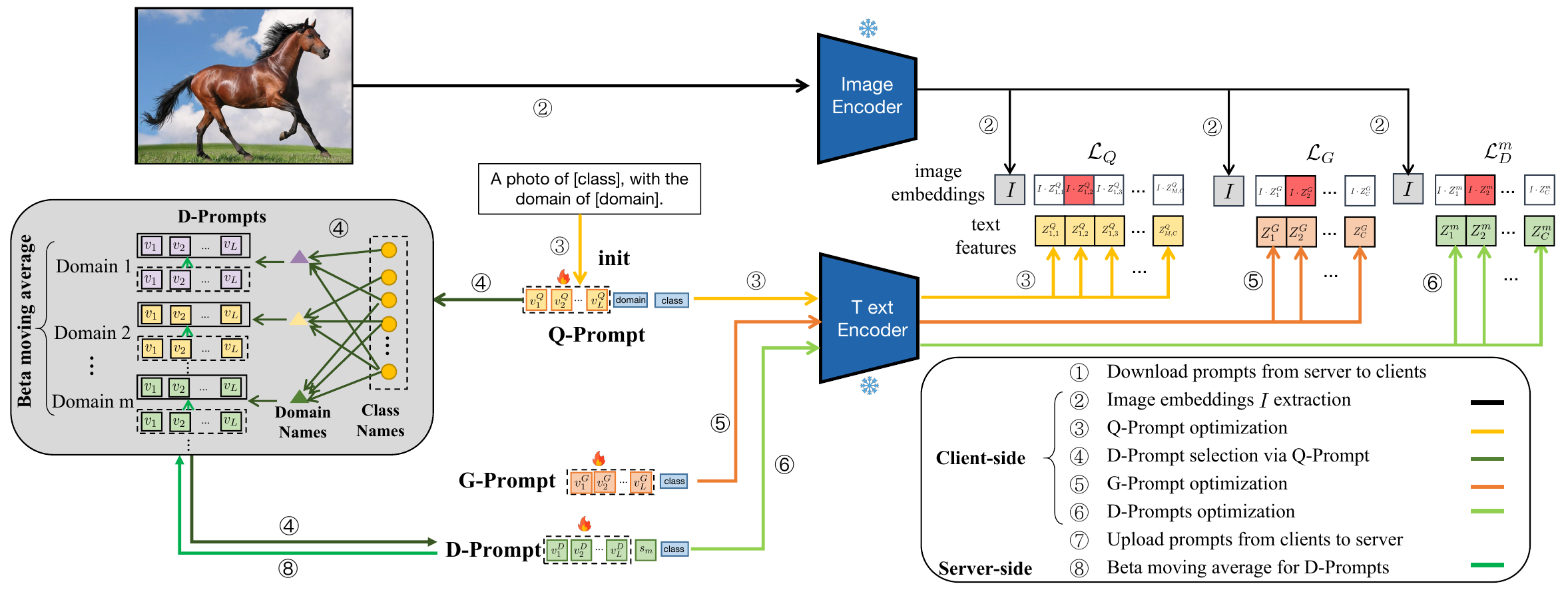}
		\hfill
	\end{center}
	\caption{
    % \zj{The font size should be bigger}
    Illustration of \textbf{Disentangled Prompt Tuning (DiPrompT) during traing}. We devise an alternative optimization strategy to update two key modules (disentangled prompt learning and dynamic query scheme), which mainly contain six steps except communication between clients and server. We first generate image embeddings and update Q-Prompt via steps 2 and 3, respectively. Then the suitable D-Prompt is selected using Q-prompt in step 4. G-Prompt and D-Prompt in disentangled prompt learning are simultaneously optimized using steps 5 and 6. Finally, we perform the beta moving average update for D-Prompts to avoid client drift in the central server in step 8.}\label{fig:main}
\end{figure*}

\section{Related Work}
\label{sec:formatting}
\textbf{Federated learning (FL)} is a decentralized training technique that leaves training data distributed on multiple decentralized clients and learns a global model by aggregating the locally uploaded parameters in a central server \citep{yang2019federated}. In FL, clients protect data privacy as raw data are always kept locally\cite{bai2023combating}. FedAvg \cite{mcmahan2017communication} is the first and most common FL work, which aggregates model updates by weighted averaging. One of the important challenges in FL is statistical heterogeneity among clients, in which each client contains different local data distribution with each other. A plethora of research has been done to tackle this problem, such as FedProx\cite{li2020federated}, MOON\cite{li2021model}, and FCCL\cite{huang2022learn}. While these FL studies have dealt with distribution heterogeneity among local clients (source domains), they ignored generalization to unseen target domains, which is the problem we mainly focus on in this paper.

\noindent \textbf{Domain generalization (DG)} generalizes a learned model from multiple source domains to unseen target domains and motivates extensive studies in a centralized setting \cite{zhou2022domain}. Representative methods either learn domain-invariant knowledge across multiple source domains \cite{nguyen2021domain, wang2021unsupervised, bai2021multi, li2018domain} or employ the idea of meta-learning \cite{du2020metalearning, li2018metalearning}. %, balaji2018metareg zhao2020domain
%, which assumes that performance improvements among source domains can increase results in unseen test domains. 
However, these DG methods require access to data from all source domains in a centralized server and ground-truth domain labels for each sample, which is usually impractical in FL scenarios due to privacy protection \cite{nguyen2022fedsr}. Moreover, some methods perform domain generalization without domain labels \cite{zhou2021domain, matsuura2020domain}, assuming the latent domain of images is reflected in their style or divided into multiple latent domains using clustering algorithms, then assigning pseudo domain labels to each sample. However, they cannot utilize text information about underlying domains. Recently, several works \cite{tenison2022fedGMA, nguyen2022fedsr, zhang2021fedADG} try to solve the DG task in the FL context. 
%FedDG \cite{liu2021feddg} is designed for medical image segmentation, which required sharing the amplitude spectrum of images among local clients. 
%FedADG \cite{zhang2021fedADG} aimed to match different distributions from source domains to a reference distribution using a generative model. 
%\cite{tenison2022fedGMA} proposed a federated gradient masking averaging (FedGMA) aggregation strategy to replace naive averaging of parameters. %in FL to improve generalization across clients and of the global model.
%Among them, FedCLIP \cite{lu2023fedclip} constructs an additional attention-based adapter in a large vision-language model.
%to achieve fast generalization.
%Nevertheless, these FL methods assume that the number of clients equals the number of domains. 
In this paper, we consider a more realistic and challenging scenario, where the number of clients is more flexible regardless of source domains, and the domain labels are unknown. %to which each sample belongs is unknown.

\noindent \textbf{Prompt Tuning} is a super efficient transfer learning paradigm \cite{liu2023promptsurvey, shin2020autoprompt},
%in natural language processing (NLP) 
whose core idea is to add little learnable embeddings at the input tokens and fast adaptation for the large language model to various downstream tasks without re-training model parameters. Early works \cite{petroni2019language, poerner2019bert} aim to manually design prompt templates based on human prior knowledge. Furthermore, CoOp \cite{zhou2022CoOp} and its extended versions \cite{chen2023plot, yao2023KCoOp, zhou2022CoCoOp} utilize a set of continuous vectors in the language branch of a large vision-language pre-trained model (CLIP), and achieve great performance improvement on multiple few-shot visual recognition tasks\cite{li2022probing, li2021groupformer, bai2021mfi}.  %Recently, PromptFL \cite{guo2022promptfl} introduces prompt tuning into FL setting to reduce resource burden in computation and communication. 
Compared with these methods, DiPrompT further extends prompt tuning into federated domain generalization by extracting multiple valuable knowledge based on prompt tuning and achieves superior generalization ability.

\section{Methodology}
In this section, we present preliminary knowledge, followed by the introduction of two key modules in training time: disentangled prompt learning and a dynamic query metric for domain queries. The overall pipeline is illustrated in Figure \ref{fig:main}, with accompanying pseudo-code in the supplementary materials. Finally, we introduce a collaborative ensemble scheme during inference.
%In this section, we first provide preliminary knowledge. Then, we propose two key modules in training time, i.e., disentangled prompt learning and a dynamic query metric concerning domain query, and their overall pipeline is shown in Figure \ref{fig:main} and we provide pseudo-code in supplementary materials. Finally, we introduce a collaborative ensemble scheme during inference.

\subsection{Preliminary and Notations}
\bsk{Federated Learning (FL) aims to utilize $K$ local models
%, denoted as ${ \mathcal{C}_1, ..., \mathcal{C}_K}$. that accurately works on test data 
to train a global model $f_g$ with parameter $\theta_g$ through $R$ global rounds, in which each randomly selected local model is trained with $T$ local iterations per global round.} Domain generalization (DG) trains a model using $M$ source domains  $D^S = \left \{D_m^S \right\}^M_{m=1}$ and is targeted to achieve decent performance on the unseen target domain $D^t$. Each source domain has a unique data distribution but shares the same task (e.g., image classification) with other domains. $m$-th domain $D_m^S$ can be represented as $ \left \{x_j, y_j, d_m\right\}^{M_m}_{j=1}$ if the domain label of each sample is known and denoted as $d_m$.

\bsk{For conventional federated DG settings, the client/domain number is limited and there exists a strict one-to-one mapping between clients and domains (i.e., $K=M$). }
%For conventional federated DG settings, the client/domain number is quite small and there exists a strict one-to-one mapping between clients and domains (i.e., $K=M$) in previous methods. 
%Since numerous edge devices in the real world can be considered as clients, this may be an impractical restriction. 
%such that satisfy the requirement of both cross-device and cross-silo settings regardless of domain diversity,
\bsk{In contrast,} we relax the above limitation into cross-device scenarios, allowing for greater flexibility in choosing the number of clients. \sk{$K$ can significantly exceed $M$ (i.e., $K \gg M$) and multiple clients may possess datasets originating from a common domain. }
%Compared with conventional FL, our setting is more challenging, where the data across clients is not only heterogeneous but from different domains rather than from the overall same distributions. 
Meanwhile, we consider a more practical requirement called latent domain generalization, where $D_m^S=\left\{x_j, y_j\right\}^{M_m}_{j=1}$ since domain label $d_m$ is not given.

\subsection{Disentangled Prompt Learning}
Although learning a single shared prompt across all source domains/clients enables extracting invariant knowledge, it may be too casual and filter out valuable information that only appeared on a subset of clients, especially when the number of clients dramatically exceeds the volume of source domains. Thus, we present disentangled prompt learning, which includes two key subcomponents: global prompt tuning and domain prompt tuning. The former aims to achieve global optimization across all clients, and the latter is utilized to extract valuable specific knowledge from different domains. We will make a detailed introduction for them as follows.

\subsubsection{Global Prompt Tuning (G-Prompt)}
To capture generic characteristics shared among all clients, G-Prompt performs global optimization using a single prompt. Concretely, each client is equipped with a pre-trained vision-language model (CLIP), which includes a frozen text encoder $g$ and visual encoder $f$. Given an image $x$ along with its label $y$, the visual embeddings $I=f(x)$ can be extracted by the visual encoder. For text embeddings, the context prompt is a set of $L$ learnable vectors $V^G = \left \{v_1^G, v_2^G, ..., v_L^G\right\}$, where the embedding dimension of each element is $d$. For the $j$-th class, the whole input for the text encoder is
$t_j=\left \{V_g, c_j \right\}$, \bsk{where $c_j$ indicates the word embedding corresponding to $j$-th class name with the same embedding dimension $d$}. Thus, we can obtain corresponding text embeddings as $Z_j^G=g(t_j)$.

%By forwarding the above image-text pairs, 
Furthermore, the local model computes the prediction probability $P(y=j|x)$ using the extracted image-text embeddings pair, which maximizes the cosine similarity score ${\rm sim}(\cdot, \cdot)$ for the correct pairs while minimizing those incorrect pairs. It can be formularized as:
\begin{equation}
P_g(y=j|x)= \frac{{\rm exp}({\rm sim} (I,Z_j)/\tau)}{\sum_{i=1}^{C}{\rm exp}({\rm sim} (I,Z_i)/\tau)},
\label{G-Prompt}
\end{equation}
where $C$ is the number of categories. We optimize the global prompt with the cross-entropy loss between the prediction probability and its label during local iterations as: 
\begin{equation}
\mathcal{L}_{G}(x,y) = \mathbb{E}_{x, j}\mathcal{L}_{ce}(y, P_g(y=j|x)).
\end{equation}
 %as initialization.
%omit valuable data from source domains when clients greatly exceed domains.

\subsubsection{Domain Prompts Tuning (D-Prompts)}
\sk{While G-Prompt acquires generic knowledge across all clients, it may overlook other valuable information from diverse clients. This is particularly evident when the number of clients significantly surpasses the domains, and diverse clients may hold data originating from a shared domain.}
%While G-Prompt can obtain generic knowledge via a single shared prompt for all clients, it tends to discard other valuable information from diversified clients. For instance, some of them are potentially useful for the target domain, especially when the number of clients significantly exceeds the domains and data from a source domain is discrete into multiple clients. 
\sk{To this end, we devise domain-specific prototypical prompts, to separately extract the distinct knowledge from the corresponding domains.} Specifically, we build a pool of domain-wise prompt pool $V^D = \left \{V_1^D,..., V_M^D \right \}$, where $V_m^D = \left \{ v_1^D, v_2^D, ..., v_L^D,s_m  \right \}$ is the prompt for $m$-th source domains and $s_m$ denotes the text embeddings corresponding to $m$-th source domain name. For $j$-th class from $m$ source domain, the text encoder produce embeddings $Z_j^m = g(t_j^m)$, where $t_j^m = \left \{ V_m^D, c_j\right\}$.
% Instead of random initialization in G-Prompt, we initialize each D-Prompt using the embedding derived from ``a photo of a [class] with the domain of [domain name]" to further enhance the domain identity for each prompt. %We hope to obtain more fine-grained knowledge from data shared in each source domain via domain prompts learning. 

To train D-Prompts in local clients, we first optimize each element using cross-entropy loss between the prediction probability of image-text pairs and the ground-truth label.
Moreover, to prevent them from progressively converging towards the same point, a contrastive loss is employed, where positive pairs involve hand-crafted prompts with the same domain names and negative pairs incorporate D-Prompts from other domains. Formally, the optimization of $m$-th domain prompt can be expressed as follows:
\begin{equation}
\begin{split}
\mathcal{L}_{D}^m(x,y) &= \mathcal{L}_{ce}^m + \mathcal{L}_{cont}^m \\
&=\mathcal{L}_{ce}^m-\log \frac{ {\rm exp}({\rm sim}( V_m^D \cdot \tilde{V}_m)))}{\sum_{i=1}^M  {\rm exp}( {\rm sim} (V_m^D\cdot V_i^D))},
\end{split}
\label{DPL_loss}
\end{equation}
where $\mathcal{L}_{ce}^m=\mathbb{E}_{x, j}\mathcal{L}_{ce}(y, P_m(y=j|x))$. and $P_m(y=j|x)=\frac{ {\rm exp}( {\rm sim}(I,Z_j^m))/\tau)}{\sum_{i=1}^{C} {\rm exp}( {\rm sim}( I,Z_i^m)/\tau)}$ denotes the prediction probability with $m$-th domain prompt for $j$-th class. Besides, $\tilde{V}_m$ indicates the hand-crafted prompt for $m$-th domain, i.e., ``a photo of a [class] with the domain of \textless domain \textgreater", where ``\textless domain \textgreater" is the text of $m$-th domain and ``[class]" denotes all potential category options. 

%\mathcal{L}_{ce}(y, P_m(y=k|x))
%
Furthermore, since multiple clients may hold data originating from a shared domain. When $K>M$, we devise a domain-wise aggregation strategy to aggregate knowledge from the same domains. \bsk{It can be represented as a weighted combination of those prompts from the same domain}: %and spread to different clients
% \begin{equation}
% \begin{aligned}
% w_{i,j}&= |t_{i,j}^{r+1}-t_j^r|/|t_{i,j}^{r+1}-t_j^r+\epsilon| \\
% t_{j}^{r+1} &= t_j^r + \frac{\sum_{i=1}^K(|\mathcal{D}_k|*w_{i,j})\cdot(t_{i,j}^{r+1}-t_j^r)}{\sum_{i=1}^K(|\mathcal{D}_k|*w_{i,j})}
% \end{aligned}
% \label{DPL_aggregation}
% \end{equation}
\begin{equation}
V_{m}^{D, {r+1}} = V_{m}^{D, {r}} + \frac{\sum_{i=1}^K(|\mathcal{D}_i|*\mathcal{I}_{m,i})\cdot \triangle V_{m,i}^{D, {r+1}}}{\sum_{i=1}^K(|\mathcal{D}_i|*\mathcal{I}_{m,i})},
\label{DPL_aggregation}
\end{equation}
\bsk{where $\triangle V_{m,i}^{D, {r+1}}=V_{m,i}^{D, {r+1}}-V_{m}^{D, {r}}$. For $m$-th domain, $V_{m, i}^{D, {r+1}}$ is uploaded prompt from $i$-th client in $r+1$-th global round, and $V_{m}^{D, {r}}$ denotes updated prompt after aggregation in $r$-th global round.
$\mathcal{I}_{m, i}$ is the output (0 or 1) of indicator function $\mathcal{I}(|\triangle V_{m,i}^{D, {r+1}}|)$ and $\mathcal{D}_i$ indicate the local data in $i$-th client. The operation only aggregates those updated prompts and filters those unchanged ones to ensure valid learning.}
%$\epsilon$ represents a small number close to 0. 
%\mathbb{E}_{x, k}\mathcal{L}_{ce}(y, P(y=k|x))

%\textbf{Updating mechanism for prompts}
%Despite multiple specific knowledge via different domain prompts, optimizing the learnable vectors in each domain prompt risks client drift. 
Despite diversified knowledge extracted from different domains, separately optimizing the domain prompt for each domain risks client drift. 
%Despite multiple specific knowledge being captured via different domain prompts, optimizing the learnable vectors in each domain prompt still faces client drift
Inspired by the pre-trained vision-language update in CLIPood \cite{shu2023clipood}, We employ a beta momentum averaging mechanism to update domain prompts. Unlike exponential moving averages that underweight initial parameters, beta momentum averaging can avoid forgetting domain information appending at the initial period. Specifically, when $m$-th domain prompt is updated from $V_m^{D,r}$ to $V_m^{D,{r+1}}$ between two adjacent rounds $r$ and $r+1$, a momentum average prompt $\hat{V}_m^{D,{r+1}}$ can be computed by:
\begin{equation}
\hat{V}_m^{D,{r+1}}= \frac{\sum_{i=0}^{r} \alpha_i}{\sum_{i=0}^{r+1} \alpha_i}\hat{V}_m^{D,r} + \frac{\alpha_r}{\sum_{i=0}^{r}}V_m^{D,{r+1}},
\label{bma}
\end{equation}
where $\alpha_i = \textbf{B}(\beta,\beta)(\frac{i+0.5}{R+1})$ and $\beta$ is a hyper-parameter for beta distribution $\textbf{B}(\cdot, \cdot)$. We set $\beta=0.2$ to preserve current optimization and initial domain knowledge attached at the beginning of training.  $R$ is the number of global rounds. 
%Overall, the loss functions for D-Prompts and G-Prompt can be defined as $L_{all}(x) = L_G(x)+ \lambda L$

After $T$ local iterations, only G-Prompt and D-Prompts are uploaded to the central server. Like FedAvg, the server updates the global prompt using a weighted aggregation of prompts from received clients in the current round. The process continues for next round by sending updated G-Prompt and momentum-averaged D-Prompts to newly selected clients.

\subsection{Dynamic Query Scheme (Q-Prompt)} %using clustering algorithms 
% Furthermore, when domain labels are not provided during training in DG, 
To efficiently learn prompts with unknown domain labels,  
we design a dynamic query scheme based on prompt tuning, which automatically selects appropriate domain prompts for different source inputs. Considering data is naturally separated according to semantic categories, most prior methods cannot avoid interference from the semantic category labels \cite{zhou2022domain}. Our query scheme adopts a two-substep strategy as shown in Figure \ref{fig:main} middle left (gray area), where each input image is first classified into a category and then grouped to an underlying domain.
Practically, we construct a learnable prompt called Q-Prompt $V^Q=\left \{v_1^Q, v_2^Q, ..., v_L^Q\right\}$ that is shared among all classes and domains. For $j$-th class and $m$-th domain, the output text embedding is $Z_{j,m}^Q =g(t_{j,m})$, where $t_{j,m}=\left\{V^Q, c_j, s_m \right\}$. We perform class and domain similarity matching on the input text-image pairs by:
\begin{equation}
P(y=j, d=m|x)= \frac{ {\rm exp}( {\rm sim}(I,Z_{j,m}^Q)/\tau)}{\sum\limits_{p=1}^{C}\sum\limits_{q=1}^{M}  {\rm exp}( {\rm sim}(I,Z_{p,q}^Q)/\tau)},
\label{Q-Prompt_match}
\end{equation}
where we select the domain with the highest probability under the ground-truth class during training and across all categories when testing on the target domain.

To ensure the effectiveness of the Q-prompt during the early period training, we initialize it with word embeddings derived from the hand-crafted prompt, i.e., ``a photo of a [class] with the domain of [domain]". %Furthermore, we utilize two self-consistency regularization terms to perform Q-prompt tuning for better adapting downstream tasks. 
Furthermore, we utilize two self-consistent regularization terms to perform Q-prompt tuning for better adaptation to downstream tasks. Specifically, MSE loss regulates feature-level alignments between the current prompt and beta momentum averaging counterpart, while we impose logits-level constraints by minimizing the Kullback-Leibler divergence between input images with current and momentum prompts, respectively. Formally, the optimization can be formalized as:
\begin{equation}
\begin{split}
\mathcal{L}_{Q} &= \mathcal{L}_{mse} +\mathcal{L}_{KL} = \mathbb{E}_{x, y}\sum_{m=1}^M[(Z_{y,m}^Q)-(\hat{Z}_{y,m}^Q)]^2\\
&+\mathcal{D}_{KL}( {\rm sim}(I,Z_{y,m}^Q),  {\rm sim}(I,\hat{Z}_{y,m}^Q))],
\end{split}
\label{QPrompt_loss}
\end{equation}
\bsk{where $Z^Q_{y,m}=g(\left\{ V^Q, c_y, s_m \right \})$ and $\hat{Z}^Q_{y,m}=g(\left\{ \hat{V}^Q, c_y, s_m \right \})$}. $\hat{V}^Q$ is the beta momentum average for Q-Prompt $V^Q$.  Given input data $(x,y)$, the overall optimization for disentangled prompt tuning is $\mathcal{L}_{}(x,y)= \mathcal{L}_G(x,y)+\lambda \mathcal{L}^m_D(x,y)$, where $\lambda$ is a weighted coefficient to balance G-Prompt and D-Prompts, and $m$ is the predicted domain via Q-Prompt. %We devise an alternative optimization strategy. %Moreover, the pseudo-code is provided in supplementary materials.

% \begin{figure}[t]
%    \centering
%    \includegraphics[width=0.99\linewidth]{./figs/fig_ensemble.pdf}
%    \caption[Caption for LOF]{ The pipeline of collaborative ensemble process in \textbf{DiPrompT in test time.} }
%    \label{ensemble}
% \end{figure}

\subsection{Collaborative Ensemble Process}
During training, we can learn general and specific knowledge from different source domains via G-Prompt and D-Prompts. In inference time, it is essential to extend this valuable knowledge %that is useful knowledge from source domains 
into the target domain. One straightforward way is to average the information from all optimized prompts. However, this method ignores the relationship (feature distance) differences between target samples and different source domains. To this end, we build a dynamic ensemble metric in our DiPrompT, which considers the above vary while effectively exploiting valuable knowledge in global prompt and various domain prompts for better target inference.  Concretely, %as shown in Figure \ref{ensemble}, 
we construct ensembled text features $Z=\left\{Z_1,..., Z_C \right\}$ for each target domain sample, which is a dynamic weighted combination of knowledge from different prompts. The ensembled text embeddings $Z_j$ for $j$-th class is defined as:
% \begin{equation}
% g^*(x) = \sum_{s=1}^M  \mathcal{I}(> \eta) g(t_s) + g(t_g)
% \label{aa}
% \end{equation}
\begin{equation}
Z_j = \sum_{m=1}^M w_m Z^m_j+ w_g Z^G_j,
\label{CSP}
\end{equation}
where we set $w_g=1$. $w_m = \frac{\max_j {\rm sim}(I\cdot Z^m_j)}{\sum_{i=1}^M \max_j I\cdot Z^i_j}$ represents the highest prediction for $m$-th domain aross all categories. 

\section{Experiments}
In this section, we conduct extensive experiments in three benchmark datasets with domain distribution shifts to demonstrate the effectiveness and robustness of our method. More details and additional experiments can be found in the supplementary material.

\begin{table*}[h]
	\begin{center}
		%\begin{spacing}{1.05}
		
% 		The best results are highlighted with bold fonts.
            
		\resizebox{0.95\hsize}{!}{
			\begin{tabular}{c| c|c|c| c| c|c| c| c|c|c|c}
				%\toprule[1.3pt]
				%\multicolumn{2}{c|}{\multirow{2}{*}
                    \hline
                    
				\multicolumn{2}{c|}{} & \multicolumn{5}{c|}{PACS} & \multicolumn{5}{c}{Officehome}\\
				\cline{3-12}
				 \multicolumn{2}{c|}{Methods} & { Art} & {Cartoon} & {Photo} & {Sketch} & {Avg} & {Art} & {Clipart} & {Product} & {Real} & {Avg}\\
      
				\hline \hline
                    \multirow{3}{*}{\shortstack{Centralized \\Learning}}& SWAD  &89.30  &83.40  &98.10  &82.60  &88.79  &66.10  &59.90  &78.70 &80.70 &70.60 \\
                   &I2ADR  &82.90  &80.80 &95.00  &83.50  &85.60  &70.30  &55.10  &80.70 &79.20 &71.40 \\
                   &  PCL &90.20  &83.90  &98.10  &82.60  &88.70  &67.30  &\textbf{59.90}  &78.70 &80.70 &71.60 \\  \hline
                   \multirow{2}{*}{FL} &FedAvg &80.41  &77.55  &92.33  &63.31  &78.39  &61.84  &51.15  &76.41 &77.39 &66.69 \\
                    &FedProx &79.19  &78.96  &94.92  &64.28  &79.33 &62.34   &52.00  &76.62  &78.56  &67.37 \\ \hline
                   \multirow{3}{*}{FL+DG} &FedADG  &78.02  &79.24  &88.50  &64.25  &77.50  &62.75  &51.43  &74.07 &77.98 &66.55 \\
                    %FedGA &45.68  &43.83  &40.34  &85.56  &84.84  &82.24  &40.73  &39.00 &28.09 &28.09 \\
                    &FedSR &82.00  &82.95  &93.53  &66.29  &81.19  &62.75  &49.85  &72.24 &74.10 &64.73 \\ 
                    &FedGMA &83.88 &80.04 &95.78  &69.17  &82.21  &65.34  &52.11  &75.98  &79.59 &68.25  \\ \hline
                    \multirow{3}{*}{\shortstack{FL+ \\Adapter/Prompt}} &PromptFL  &92.77  &94.24  &99.40  &82.13  &92.13  &72.18  &56.88  &82.13 &84.78 &73.99 \\
                    &FedCLIP &92.93   &94.80  &99.52  &82.26 &92.37  &71.03  &56.13  &83.76 &84.14 &73.76 \\
                    &\textbf{Ours} &\textbf{94.97} &\textbf{96.25} &\textbf{99.56 } &\textbf{84.72} &\textbf{93.88}&\textbf{74.21} &58.90 &\textbf{85.51} &\textbf{86.12}  &\textbf{76.18} \\ \hline 	
			\end{tabular}
		}
  \caption{Performance comparison of our proposed DiPrompT with state-of-the-art methods on PACS and Officehome datasets. FedDure outperforms all other methods}\label{pacs_office_comparasion}
		%\end{spacing}
	\end{center}
 
\end{table*}
%\endgroup

\begin{table}[h]
	\begin{center}
		%\begin{spacing}{1.05}
		
% 		The best results are highlighted with bold fonts.
            
		\resizebox{1.0\hsize}{!}{
			\begin{tabular}{c|c|c| c| c|c}
				%\toprule[1.3pt]
				%\multicolumn{2}{c|}{\multirow{2}{*}
                    \hline
				\multirow{2}{*}{Methods}  & \multicolumn{5}{c}{VLCS}\\
				\cline{2-6} 
				  & {Caltech} & {Labelme} & {Pascal} & {Sun} & {Avg}\\
				\hline \hline
                    Fishr &98.90  &64.0  &71.50  &76.80  &77.80  \\ 
                    ITL-NET &98.30  &65.40 &75.10  &76.80  &78.90  \\
                    VAUE &99.00  &64.70  &75.10  &79.40  &79.40   \\ \hline
                    FedAvg   &93.54  &60.69  &72.22 &74.66 &75.27 \\
                    FedProx   &94.55 &61.55  &73.75 &75.52 &76.34 \\ \hline
                    FedADG  &95.96  &61.43  &66.30 &72.44 &74.03 \\
                    FedSR   &96.37  &60.15  &69.74 &73.40 &74.91 \\ 
                    FedGMA   &97.88  &61.47  &73.76 &76.19 &77.32 \\ \hline
                    PromptFL   &99.49  &65.36  &78.54 &78.75 &80.53 \\
                    FedCLIP   &99.39  &66.70  &82.24 &78.62 &81.73  \\ 
                    \textbf{Ours} &\textbf{99.70 } &\textbf{69.23 } &\textbf{84.16 } &\textbf{81.72} &\textbf{83.70} \\ \hline 	
			\end{tabular}
		}
  \caption{Performance comparison of our proposed DiPrompT with state-of-the-art methods on VLCS dataset.}\label{vlcs_comparasion}
		%\end{spacing}
	\end{center}
 
\end{table}

\subsection{Experimental Setup}
\subsubsection{Datasets.} We perform comprehensive experiments on three widely used datasets in domain generalization tasks, including PACS (4 domains: photo, art-painting, cartoon, and sketch) \cite{li2017deeper}, Officehome (4 domains: Art, Clipart, Product, and Real World) \cite{venkateswara2017deep}, and VLCS (4 domains: Pascal, Labelme, Caltech and Sun) \cite{fang2013unbiased}. To conduct our analysis on each dataset, we adopt the "leave-one-domain-out" strategy, wherein we select one domain as the target domain and utilize the remaining domains as source domains. We provide detailed dataset descriptions in supplementary materials.
% Comprehensive details about the datasets are available in the supplementary material.
%For each dataset, we employ the ``leave-one-domain-out" strategy, where we choose one domain to serve as the target domain and use the rest domains as source domains. We provide detailed dataset descriptions in supplementary materials.

\subsubsection{Baselines.}
We compare our DiPrompT with the following state-of-the-art methods. SWAD \cite{cha2022swad}, I2ADR\cite{meng2022i2ADR}, PCL \cite{yao2022pcl}, Fishr \cite{rame2022fishr}, ITL-Net \cite{gao2022ITLNet} and VAUE \cite{lin2022mitigating}  are sota centralized learning methods that learned a generalized model by using all sources domains in a data pool regardless of data privacy. Meanwhile, we consider recent FL algorithms as our main competitors. FedAvg and FedProx are common FL algorithms that aimed to tackle data heterogeneity between clients.  FedADG\cite{zhang2021fedADG}, FedSR\cite{nguyen2022fedsr}, and FedGMA\cite{tenison2022fedGMA} focused on solving the Federated domain generalization (FL+DG). PromptFL \cite{guo2022promptfl} and FedCLIP \cite{lu2023fedclip} adapted pre-trained vision-language models in FL tasks via prompt and adapter tuning, respectively.

\begin{figure*}
\begin{subfigure}{0.245\textwidth}
\includegraphics[width=\textwidth,]{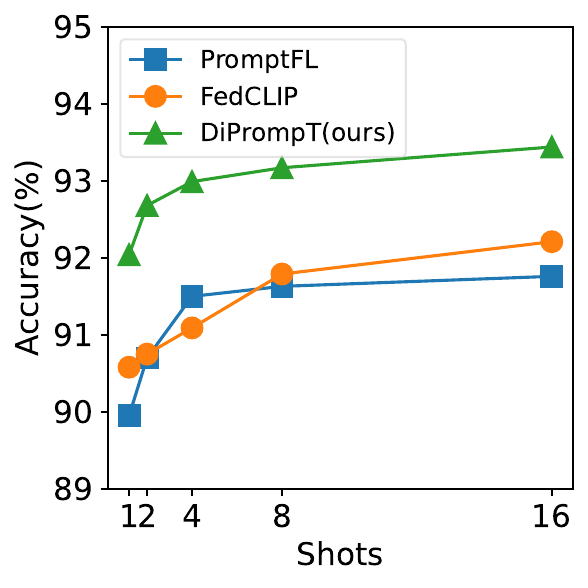} 
\caption{PACS} \label{shots_pacs}
\end{subfigure}
\begin{subfigure}{0.245\textwidth}
\includegraphics[width=\textwidth]{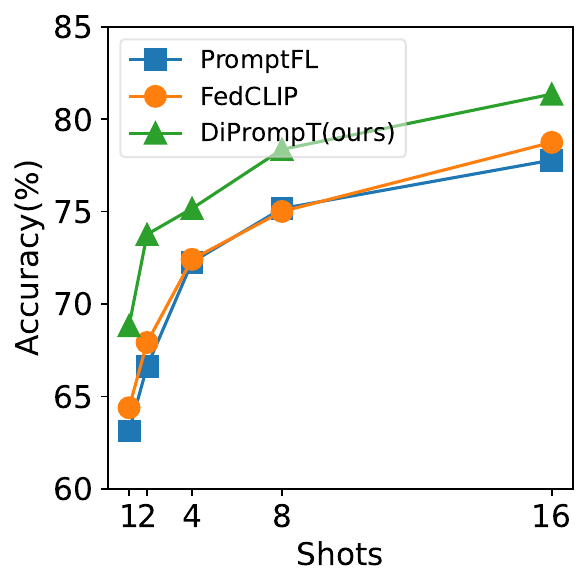}
\caption{VLCS}\label{shots_vlcs}
\end{subfigure}
\begin{subfigure}{0.245\textwidth}
\includegraphics[width=\textwidth,]{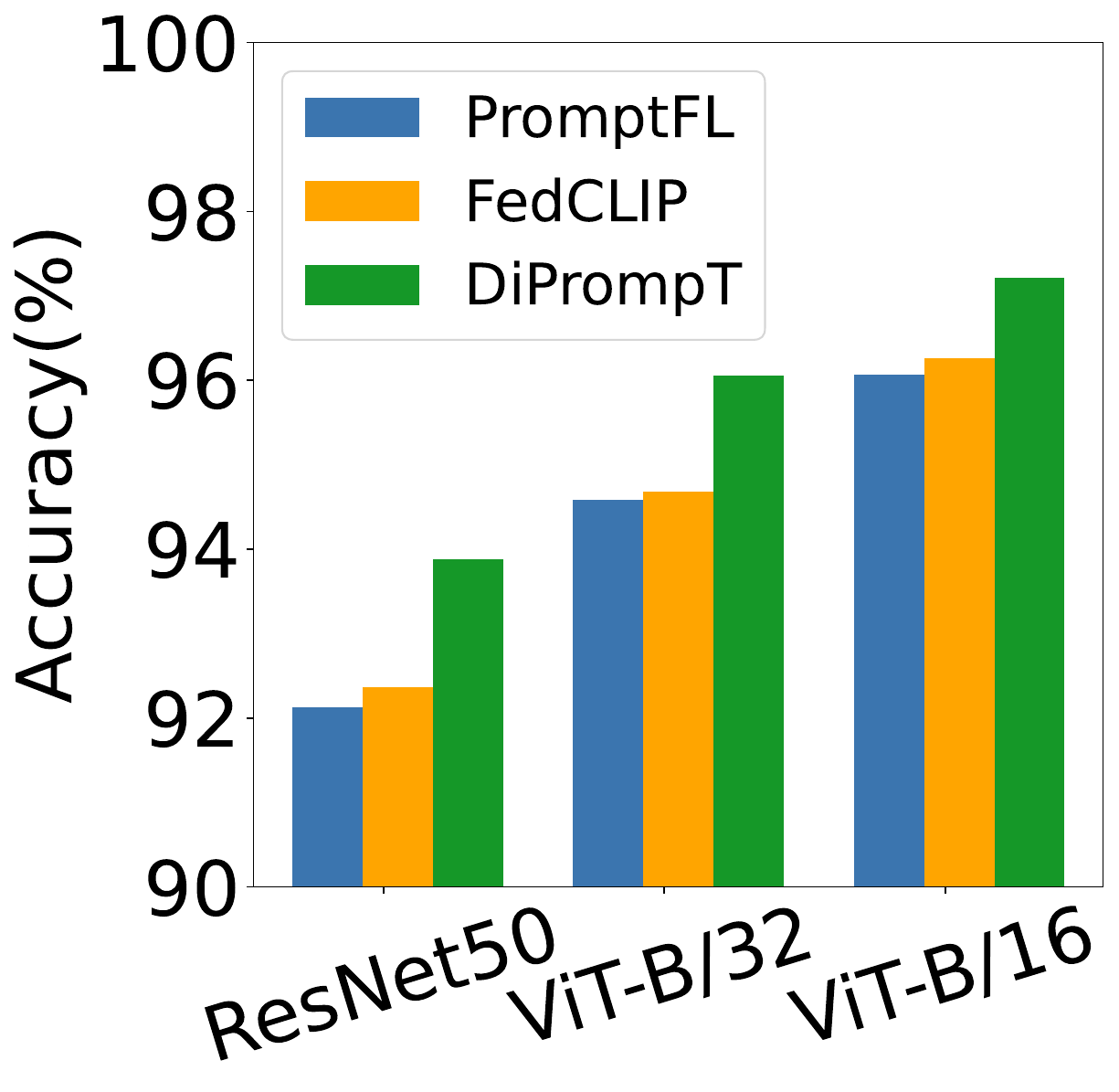} 
\caption{PACS} \label{backbone_pacs}
\end{subfigure}
\begin{subfigure}{0.24\textwidth}
\includegraphics[width=\textwidth,]{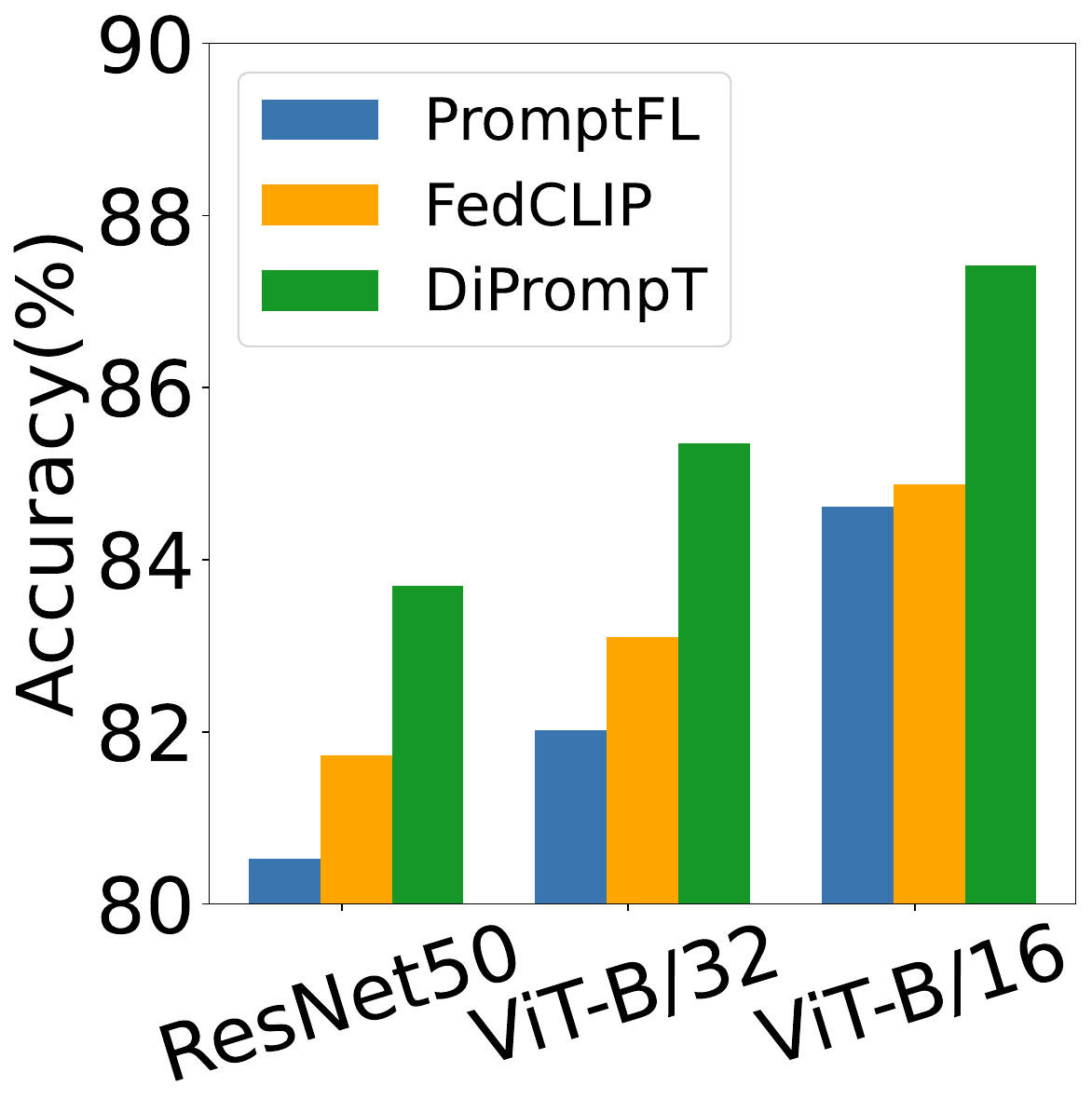} 
\caption{VLCS} \label{backbone_vlcs}
\end{subfigure}
\caption{Analysis in terms of few shots settings and different backbone architectures on PACS and VLCS datasets.}
\label{fig_dist}
\end{figure*}

\subsubsection{Implementation Details.}
In our experiment, we conducted model training using PyTorch on GeForce RTX 3090 GPU. The training process for all models involved the utilization of the Adam optimizer with a learning rate set at 5e-4. It's worth noting that this uniform configuration applied to all methods under consideration, with the exception of FedGMA. For FedGMA, we tailored its optimizer parameters to align with its best-reported hyperparameters
%we train all models with PyTorch on GeForce RTX 3090 GPU. For model training, we employ Adam optimizer with 5e-4 learning rate for all methods except FedGMA, where its optimizer followed the best-reported hyperparameters 
in \cite{tenison2022fedGMA}, ensuring a comprehensive and consistent approach across our experimental design. Furthermore, our default settings also include the weighted coefficient $\lambda = 1.0$, batch size $b=16$, local iterations $T=1$, the selected clients in each round $H=5$, the number of clients $K=20$, and the number of global rounds $R=100$. Following CLIP \cite{radford2021learning}, we adopt ResNet50 as the backbone architecture of an image encoder and use a masked self-attention Transformer as a text encoder.  Note that if there is no specified description, we use the same hyper-parameters and backbone architecture for DiPrompT and other methods in all experiments to implement fair comparison.

\begin{table}[t]
	\begin{center}
		%\begin{spacing}{1.05}
		
% 		The best results are highlighted with bold fonts.
            
		\resizebox{1.0\hsize}{!}{
			\begin{tabular}{c|c|c| c| c|c}
				%\toprule[1.3pt]
				%\multicolumn{2}{c|}{\multirow{2}{*}
                    \hline
				\multirow{2}{*}{Ablated components}  & \multicolumn{5}{c}{PACS}\\
				\cline{2-6} 
				  & {Art} & {Cartoon} & {Photo} & {Sketch} & {Avg}\\
				\hline \hline
                    %Baseline &90.31  &92.79  &99.44 &78.35  &90.22  \\ 
                    w/o G-Prompt &88.87  &91.42  &98.22  &75.47  &88.50   \\
                    w/o D-Prompts &92.77  &94.24 &99.40  &82.13 &92.13  \\
                     w/o $L_{cont}$  &94.57 &95.61  &99.52 &84.27 &93.49  \\
                    w/o Q-Prompt   &93.35 &93.69  &99.44  &83.16 &92.40 \\
                    w/o ensemble &94.19 &94.88 &99.52 &83.4      &93.00 \\
                    w/o $L_{KL}$ &94.04 &95.56  &99.40  &81.19  &92.54   \\
                    w/o $L_{mse}$ &93.75 &95.01  &99.34  &80.80  &92.23   \\
                    w domain  &95.82 &96.81 &99.64 &84.80    &94.26 \\
                    \textbf{Ours} &94.47 &96.25 &99.56 &84.72 &93.88 \\ \hline 	
			\end{tabular}
		}
  \caption{Quantitative analysis of components of DiPrompT.}\label{components_comparasion}
		%\end{spacing}
	\end{center}
\end{table}

% on PACS dataset

\subsection{Performance Comparison}
We report the experimental results of DiPrompT and other state-of-the-art methods in Table \ref{pacs_office_comparasion} and Table \ref{vlcs_comparasion}, which involves 12 domain generalization tasks on the PACS, OfficeHome, and VLCS datasets. All results are the average of three runs, and bold text represents the best results. Specifically, as mentioned earlier, we considered a more flexible scenario when $K>M$, thus there is no one-to-one mapping between local clients and domains. 
It can be observed that the previous federated domain generalization (FL+DG) methods perform poorly, even worse than the FL methods designed for client heterogeneity. FedGMA's performance is comparable to FedAvg and FedProx, while FedADG and FedSR exhibit lower performance in many DG tasks compared to FedAvg and FedProx. The phenomenon might stem from FL+DG methods struggling with valuable knowledge as client volume significantly surpasses the number of source domains.
% Actually, they are not direct competitors to our model due to the different setups (FL vs. centralized learning).
% in centralized settings

Furthermore, we compare our DiPrompT with state-of-the-art methods from centralized learning, FL, and FL+DG.  It is noted that due to the dependencies on domain labels and interrelationships between source domains, centralized learning methods cannot be applied to this domain-separated setting. They are not direct competitors to our model due to the different setups (FL vs. centralized learning), and we adopt the results taken from their original papers. With prompt tuning from the textual branch, DiPrompT outperforms centralized methods as well as FL and FL+DG techniques by a big margin, where we adopt the same image backbone (i.e. ResNet50) even when domain labels are unknown for DiPrompT. Moreover, due to complementary knowledge from G-Prompt and D-Prompts, our DiPrompT shows significant superiority over PromptFL and FedCLIP, which utilize the same pre-trained vision-language model (CLIP) as our methods. Overall, DiPrompT achieves the best average accuracy across four benchmark datasets. When examining the results for each target domain setting within each dataset, DiPrompT outperforms previous methods in 11 out of 12 settings. These quantitative results demonstrate the effectiveness of our approach.

\subsection{Ablation Study}
\subsubsection{Effectiveness of Components.}
To measure the importance of each component in our DiPrompT, we conduct an ablation analysis for the following variants using the PACS dataset in Table \ref{components_comparasion}.

Row 1 denotes that DiPrompT removes G-Prompts in each local model.  Row 2 indicates the variant that eliminates D-Prompts in our DiPrompT, while only updating G-Prompt, which is essentially equivalent to PromptFL. Compared with DiPrompT, the dramatic performance drop shows that the two components are both effective and can provide value complementary knowledge by ensembling them.
In Row 3, we remove the built contrastive loss during D-Prompts optimization and find a slight drop. 
Instead of learnable Q-Prompt, we use static prompts in Row 4 (i.e., ``a photo of a [CLASS] with the domain of [Domain].") to look up domain prompts for each sample. As results show, learnable Q-Prompt plays an important role in exploring domain information in latent domain learning.
In Row 5, we only choose the domain prompt with the highest weight rather than ensembling knowledge from different source domains during inference. The performance decrease suggests that the collaborative ensemble metric enables the sufficient exploitation of more valuable knowledge. 
Moreover, the results in Row 6 and 7 implicitly showcase the benefit of $L_{KL}$ and $L_{mse}$ for Q-Prompt optimization.
When the domain labels are given for all local training data, we can observe that our DiPrompT is close to its upper bound.  
These evaluations verify the effectiveness of each component, and DiPrompT can obtain complementary knowledge and eliminate the dependency on the domain label for better generalization on the unseen target domain.

\begin{figure}
\begin{subfigure}{0.22\textwidth}
\includegraphics[width=\textwidth,]{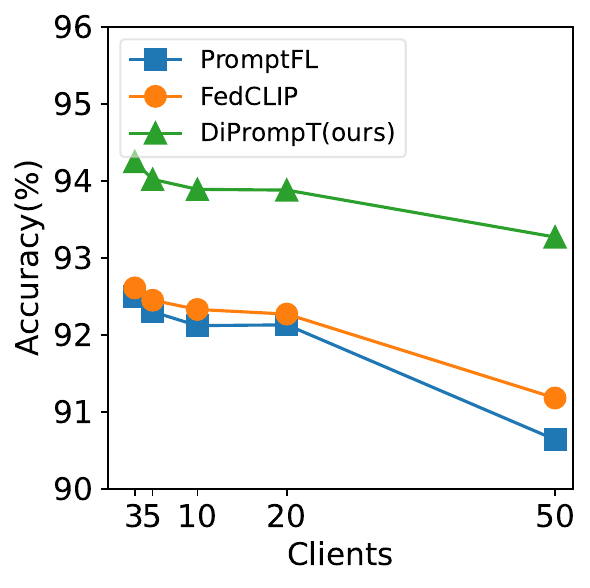} 
\caption{The number of Client $K$} \label{client_num}
\end{subfigure}
\begin{subfigure}{0.22\textwidth}
\includegraphics[width=\textwidth]{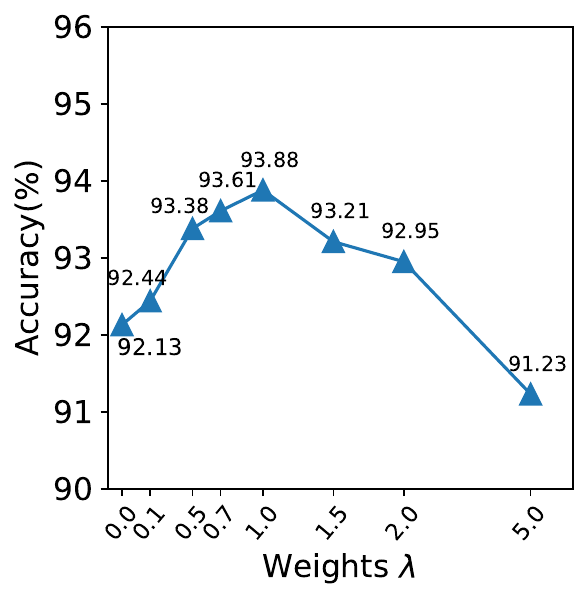}
\caption{ Weight coefficient $\lambda$}
\label{lambda}
\end{subfigure}
\caption{Hyperparameters analysis in terms of the number of clients $K$ and weight coefficient $\lambda$ on PACS.}
\label{hyperparameter}
\end{figure}

\begin{figure}
\begin{subfigure}{0.235\textwidth}
\includegraphics[width=\textwidth,]{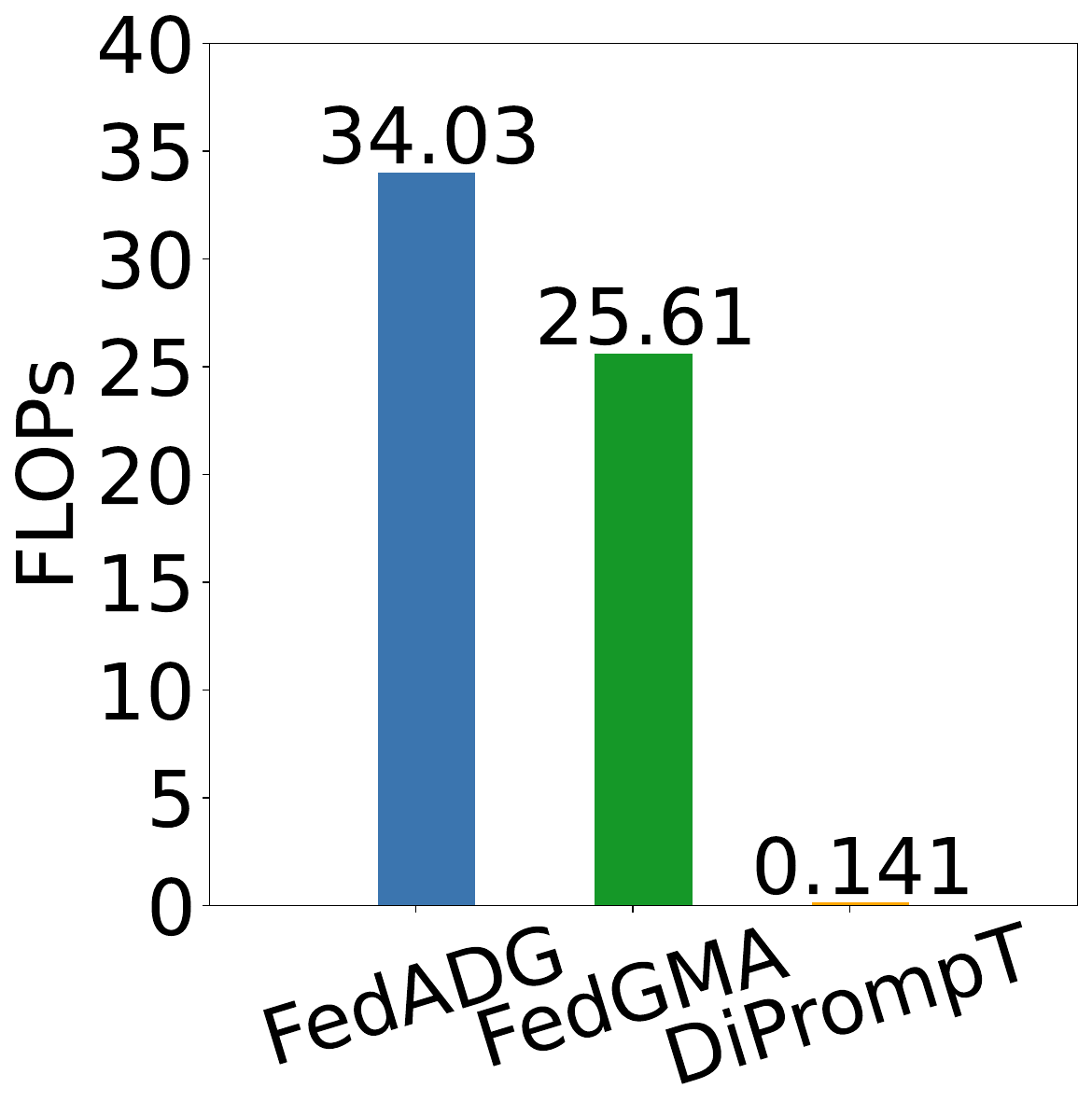} 
\caption{Communication cost} \label{communication}
\end{subfigure}
\begin{subfigure}{0.235\textwidth}
\includegraphics[width=\textwidth]{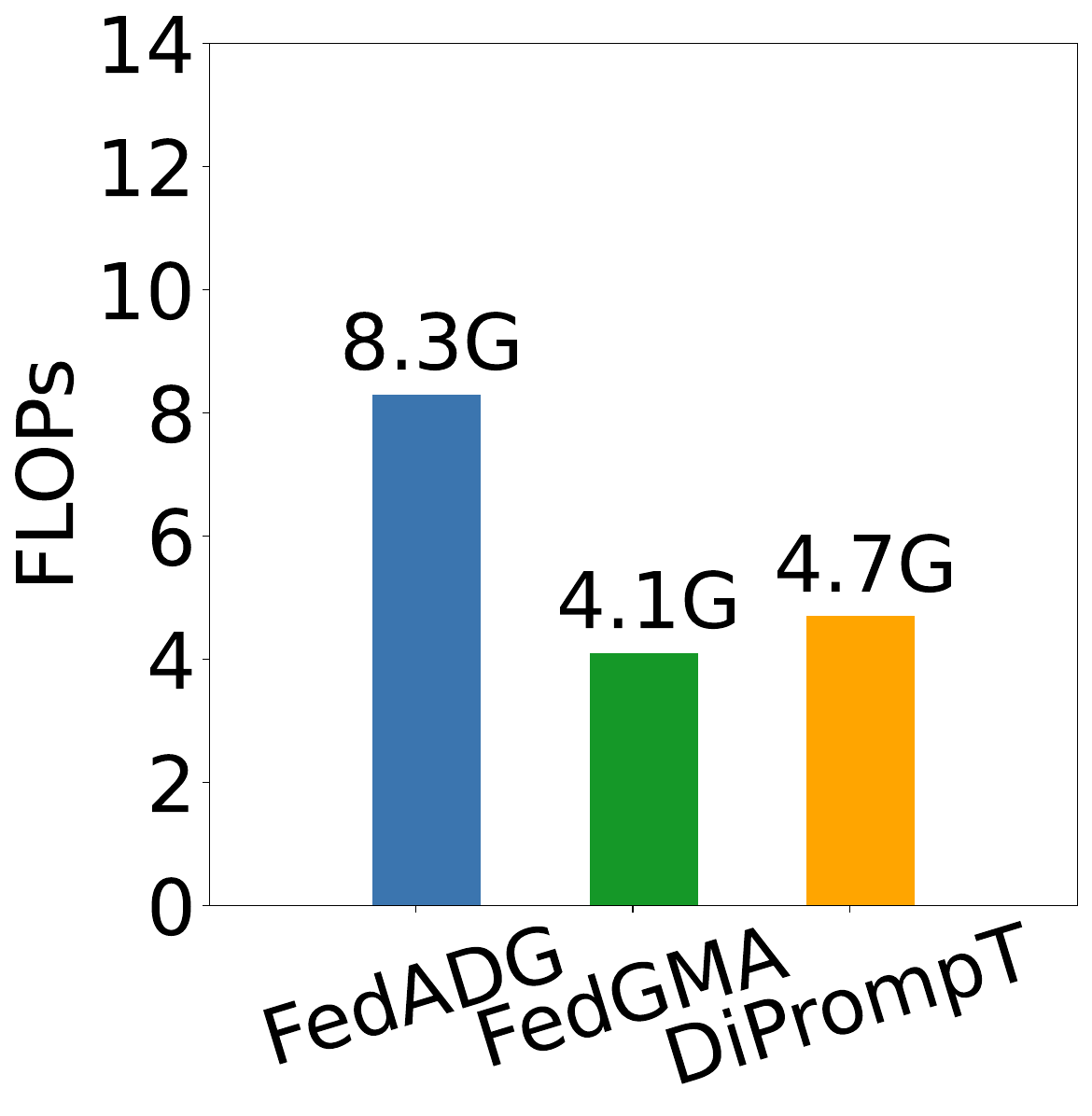}
\caption{ Computation cost}
\label{computation}
\end{subfigure}
\caption{Comparison of computation and communication cost of DiPrompT and other federated domain generalization methods.}
\label{costs}
\end{figure}

%Vit16， ViT32直方图
%fewshots 折线图
%Clients 折线图

\subsubsection{Results on Few-shot Recognition.}
To explore the effectiveness of our framework with extremely limited data, we compare the generalization performance of DiPrompT with some state-of-the-art methods across various few-shot settings (i.e., 1, 2, 4, 8, 16 shots). As depicted in Figure \ref{shots_pacs} and shots \ref{shots_vlcs}, we observe that as the number of training samples per class on each client increases, the performance of all methods is enhanced. Meanwhile, DiPrompT achieves significant performance improvements across all shot settings on PACS and VLCS datasets, and shows higher gains in minimal data cases (e.g. 1, 2 shots). Thereby, these results demonstrate the robustness of DiPrompT in acquiring complementary knowledge in data-scarce scenarios. %Detailed results for each target domain in each dataset are provided in supplementary materials.
%is capable of achieving excellent effects a
\subsubsection{Impacts of Backbone Architecture.} 
To investigate how the choice of backbone architecture impacts generalization performance, we summarize the generalization effect of multiple methods based on different image encoders, including Reset50, ViT-B/32, and ViT-B/16. As shown in Figure \ref{backbone_pacs} and \ref{backbone_vlcs},  the more advanced the backbone architecture is, the better the performance of all methods can achieve. More importantly, our DiPrompT is significantly better than other state-of-the-art methods across all backbone on PACS and VLCS datasets, which show the robustness of our DiPrompT in terms of backbones.

\subsubsection{Effect of Hyperparameters.}
There are two key hyperparameters, including the number of clients $K$ and the value of the weight $\lambda$ between G-Prompt and D-Prompts. We first investigate the performance impact of the number of total clients, which varies in $\left\{ 3,5,10,20, 50\right\}$. Note that we assign a source domain to a single client when $K=3$ and randomly select 5 clients per round when $K>=5$. As illustrated in Figure \ref{client_num}, although there is inevitable performance degradation for all methods by increasing total clients, our DiPrompT only has a light performance drop (about 1\%) and outperforms other methods across all numerical settings. This phenomenon demonstrates the challenges posed by a large number of clients in federated domain generalization as well as the effectiveness of DiPrompT. Finally, we examine the impacts of the different values of the weight hyperparameter $\lambda$ between G-Prompt and D-Prompts on performance. As shown in Figure \ref{lambda}, we observe that the optimal performance is achieved when setting  $\lambda=1$. 

%Computation and Communication Cost Analysis
\subsubsection{Cost Analysis.} \bsk{Finally, we analyze the efficiency of DiPrompT in terms of the computation and communication cost in Figure \ref{costs}. We measure the communication cost by the size of uploaded data per round. It can be observed that DiPrompT can save at most 830 times communication cost per round compared to FedADG and 625 times for FedGM, indicating that DiPrompT can significantly reduce communication burden. For the computation cost, we report the comparison of GPU time as in the same mini-batch, where DiPrompT outperforms FedADG around 2 times and is comparable computation cost with FedGMA. It is noted that we don't compare with other FL and FL+DG methods (i.e FedAvg, FedProx, and FedSR) since they have almost the same communication and computation costs as FedGMA, while PrompFL and FedCLIP have similar costs with our DiPrompT. These results demonstrate the efficiency of our DiPrompT, which can be applied to many real-world scenarios.}

\section{Conclusion}
In this work, we propose a novel framework named DiPrompT, the first attempt to introduce prompt tuning to federated domain generalization. Specifically, DiPrompT aims to learn general knowledge as well as valuable specific information across all clients, especially when the number of clients and source domains is inconsistent during training and different clients may store data originating from a shared domain. It provides more complementary knowledge for unseen target prediction during inference. Moreover, we build an adaptive query mechanism based on prompt tuning, which automatically searches the suitable domain for each sample when domain labels are not given for all local data. Extensive experiments show that our DiPrompT outperforms state-of-the-art methods, and is even better than many centralized learning strategies using domain labels.
% \section{Final copy}

% You must include your signed IEEE copyright release form when you submit your finished paper.
% We MUST have this form before your paper can be published in the proceedings.

% Please direct any questions to the production editor in charge of these proceedings at the IEEE Computer Society Press:
% \url{https://www.computer.org/about/contact}.
\newpage
{
    \small
    \bibliographystyle{ieeenat_fullname}
    \bibliography{main}

\begin{thebibliography}{45}
\providecommand{\natexlab}[1]{#1}
\providecommand{\url}[1]{\texttt{#1}}
\expandafter\ifx\csname urlstyle\endcsname\relax
  \providecommand{\doi}[1]{doi: #1}\else
  \providecommand{\doi}{doi: \begingroup \urlstyle{rm}\Url}\fi

\bibitem[Bai et~al.(2023)Bai, Bagchi, and Inouye]{bai2023benchmarking}
Ruqi Bai, Saurabh Bagchi, and David~I Inouye.
\newblock Benchmarking algorithms for federated domain generalization.
\newblock \emph{arXiv preprint arXiv:2307.04942}, 2023.

\bibitem[Bai et~al.(2021{\natexlab{a}})Bai, Gao, Wang, and Li]{bai2021multi}
Sikai Bai, Junyu Gao, Qi Wang, and Xuelong Li.
\newblock Multi-domain synchronous refinement network for unsupervised cross-domain person re-identification.
\newblock In \emph{2021 IEEE International Conference on Multimedia and Expo (ICME)}, pages 1--6, 2021{\natexlab{a}}.

\bibitem[Bai et~al.(2021{\natexlab{b}})Bai, Wang, and Li]{bai2021mfi}
Sikai Bai, Qi Wang, and Xuelong Li.
\newblock Mfi: Multi-range feature interchange for video action recognition.
\newblock In \emph{2020 25th International Conference on Pattern Recognition (ICPR)}, pages 6664--6671. IEEE, 2021{\natexlab{b}}.

\bibitem[Bai et~al.(2024)Bai, Li, Zhuang, Yang, Hou, Yi, Zhang, Gao, Zhang, and Guo]{bai2023combating}
Sikai Bai, Shuaicheng Li, Weiming Zhuang, Kunlin Yang, Jun Hou, Shuai Yi, Shuai Zhang, Junyu Gao, Jie Zhang, and Song Guo.
\newblock Combating data imbalances in federated semi-supervised learning with dual regulators.
\newblock 2024.

\bibitem[Cha et~al.(2022)Cha, Lee, Park, and Chun]{cha2022swad}
Junbum Cha, Kyungjae Lee, Sungrae Park, and Sanghyuk Chun.
\newblock Domain generalization by mutual-information regularization with pre-trained models.
\newblock In \emph{European Conference on Computer Vision}, pages 440--457. Springer, 2022.

\bibitem[Chen et~al.(2023)Chen, Yao, Song, Li, Rao, and Zhang]{chen2023plot}
Guangyi Chen, Weiran Yao, Xiangchen Song, Xinyue Li, Yongming Rao, and Kun Zhang.
\newblock Prompt learning with optimal transport for vision-language models.
\newblock 2023.

\bibitem[Du et~al.(2020)Du, Xu, Xiong, Qiu, Zhen, Snoek, and Shao]{du2020metalearning}
Yingjun Du, Jun Xu, Huan Xiong, Qiang Qiu, Xiantong Zhen, Cees~GM Snoek, and Ling Shao.
\newblock Learning to learn with variational information bottleneck for domain generalization.
\newblock In \emph{Computer Vision--ECCV 2020: 16th European Conference, Glasgow, UK, August 23--28, 2020, Proceedings, Part X 16}, pages 200--216, 2020.

\bibitem[Fang et~al.(2013)Fang, Xu, and Rockmore]{fang2013unbiased}
Chen Fang, Ye Xu, and Daniel~N Rockmore.
\newblock Unbiased metric learning: On the utilization of multiple datasets and web images for softening bias.
\newblock In \emph{Proceedings of the IEEE International Conference on Computer Vision}, pages 1657--1664, 2013.

\bibitem[Gao et~al.(2022)Gao, Gouk, Yang, and Hospedales]{gao2022ITLNet}
Boyan Gao, Henry Gouk, Yongxin Yang, and Timothy Hospedales.
\newblock Loss function learning for domain generalization by implicit gradient.
\newblock In \emph{International Conference on Machine Learning}, pages 7002--7016, 2022.

\bibitem[Guo et~al.(2022)Guo, Guo, Wang, and Xu]{guo2022promptfl}
Tao Guo, Song Guo, Junxiao Wang, and Wenchao Xu.
\newblock Promptfl: Let federated participants cooperatively learn prompts instead of models--federated learning in age of foundation model.
\newblock \emph{arXiv preprint arXiv:2208.11625}, 2022.

\bibitem[Huang et~al.(2022)Huang, Ye, and Du]{huang2022learn}
Wenke Huang, Mang Ye, and Bo Du.
\newblock Learn from others and be yourself in heterogeneous federated learning.
\newblock In \emph{Proceedings of the IEEE/CVF Conference on Computer Vision and Pattern Recognition}, pages 10143--10153, 2022.

\bibitem[Li et~al.(2017)Li, Yang, Song, and Hospedales]{li2017deeper}
Da Li, Yongxin Yang, Yi-Zhe Song, and Timothy~M Hospedales.
\newblock Deeper, broader and artier domain generalization.
\newblock In \emph{Proceedings of the IEEE international conference on computer vision}, pages 5542--5550, 2017.

\bibitem[Li et~al.(2018{\natexlab{a}})Li, Yang, Song, and Hospedales]{li2018metalearning}
Da Li, Yongxin Yang, Yi-Zhe Song, and Timothy Hospedales.
\newblock Learning to generalize: Meta-learning for domain generalization.
\newblock In \emph{Proceedings of the AAAI conference on artificial intelligence}, 2018{\natexlab{a}}.

\bibitem[Li et~al.(2021{\natexlab{a}})Li, He, and Song]{li2021model}
Qinbin Li, Bingsheng He, and Dawn Song.
\newblock Model-contrastive federated learning.
\newblock In \emph{Proceedings of the IEEE/CVF conference on computer vision and pattern recognition}, pages 10713--10722, 2021{\natexlab{a}}.

\bibitem[Li et~al.(2021{\natexlab{b}})Li, Cao, Liu, Yang, Liu, Hou, and Yi]{li2021groupformer}
Shuaicheng Li, Qianggang Cao, Lingbo Liu, Kunlin Yang, Shinan Liu, Jun Hou, and Shuai Yi.
\newblock Groupformer: Group activity recognition with clustered spatial-temporal transformer.
\newblock In \emph{Proceedings of the IEEE/CVF International Conference on Computer Vision}, pages 13668--13677, 2021{\natexlab{b}}.

\bibitem[Li et~al.(2022)Li, Zhang, Yang, Liu, Liu, Hou, and Yi]{li2022probing}
Shuaicheng Li, Feng Zhang, Kunlin Yang, Lingbo Liu, Shinan Liu, Jun Hou, and Shuai Yi.
\newblock Probing visual-audio representation for video highlight detection via hard-pairs guided contrastive learning.
\newblock \emph{arXiv preprint arXiv:2206.10157}, 2022.

\bibitem[Li et~al.(2020)Li, Sahu, Zaheer, Sanjabi, Talwalkar, and Smith]{li2020federated}
Tian Li, Anit~Kumar Sahu, Manzil Zaheer, Maziar Sanjabi, Ameet Talwalkar, and Virginia Smith.
\newblock Federated optimization in heterogeneous networks.
\newblock \emph{Proceedings of Machine learning and systems}, 2:\penalty0 429--450, 2020.

\bibitem[Li et~al.(2018{\natexlab{b}})Li, Gong, Tian, Liu, and Tao]{li2018domain}
Ya Li, Mingming Gong, Xinmei Tian, Tongliang Liu, and Dacheng Tao.
\newblock Domain generalization via conditional invariant representations.
\newblock In \emph{Proceedings of the AAAI conference on artificial intelligence}, 2018{\natexlab{b}}.

\bibitem[Lin et~al.(2022)Lin, Tang, Wang, and Zhang]{lin2022mitigating}
Jianxin Lin, Yongqiang Tang, Junping Wang, and Wensheng Zhang.
\newblock Mitigating both covariate and conditional shift for domain generalization.
\newblock In \emph{2022 IEEE 8th International Conference on Cloud Computing and Intelligent Systems (CCIS)}, pages 437--443, 2022.

\bibitem[Liu et~al.(2023)Liu, Yuan, Fu, Jiang, Hayashi, and Neubig]{liu2023promptsurvey}
Pengfei Liu, Weizhe Yuan, Jinlan Fu, Zhengbao Jiang, Hiroaki Hayashi, and Graham Neubig.
\newblock Pre-train, prompt, and predict: A systematic survey of prompting methods in natural language processing.
\newblock \emph{ACM Computing Surveys}, 55\penalty0 (9):\penalty0 1--35, 2023.

\bibitem[Liu et~al.(2021)Liu, Chen, Qin, Dou, and Heng]{liu2021feddg}
Quande Liu, Cheng Chen, Jing Qin, Qi Dou, and Pheng-Ann Heng.
\newblock Feddg: Federated domain generalization on medical image segmentation via episodic learning in continuous frequency space.
\newblock In \emph{Proceedings of the IEEE/CVF Conference on Computer Vision and Pattern Recognition}, pages 1013--1023, 2021.

\bibitem[Lu et~al.(2023)Lu, Hu, Wang, and Xie]{lu2023fedclip}
Wang Lu, Xixu Hu, Jindong Wang, and Xing Xie.
\newblock Fedclip: Fast generalization and personalization for clip in federated learning.
\newblock \emph{arXiv preprint arXiv:2302.13485}, 2023.

\bibitem[Matsuura and Harada(2020)]{matsuura2020domain}
Toshihiko Matsuura and Tatsuya Harada.
\newblock Domain generalization using a mixture of multiple latent domains.
\newblock In \emph{Proceedings of the AAAI Conference on Artificial Intelligence}, pages 11749--11756, 2020.

\bibitem[McMahan et~al.(2017)McMahan, Moore, Ramage, Hampson, and y~Arcas]{mcmahan2017communication}
Brendan McMahan, Eider Moore, Daniel Ramage, Seth Hampson, and Blaise~Aguera y Arcas.
\newblock Communication-efficient learning of deep networks from decentralized data.
\newblock In \emph{Artificial intelligence and statistics}, pages 1273--1282. PMLR, 2017.

\bibitem[Meng et~al.(2022)Meng, Li, Chen, Yang, Song, Wang, Zhang, Song, Xie, and Pu]{meng2022i2ADR}
Rang Meng, Xianfeng Li, Weijie Chen, Shicai Yang, Jie Song, Xinchao Wang, Lei Zhang, Mingli Song, Di Xie, and Shiliang Pu.
\newblock Attention diversification for domain generalization.
\newblock In \emph{European conference on computer vision}, pages 322--340. Springer, 2022.

\bibitem[Nguyen et~al.(2021)Nguyen, Tran, Gal, and Baydin]{nguyen2021domain}
A~Tuan Nguyen, Toan Tran, Yarin Gal, and Atilim~Gunes Baydin.
\newblock Domain invariant representation learning with domain density transformations.
\newblock \emph{Advances in Neural Information Processing Systems}, 34:\penalty0 5264--5275, 2021.

\bibitem[Nguyen et~al.(2022)Nguyen, Torr, and Lim]{nguyen2022fedsr}
A~Tuan Nguyen, Philip Torr, and Ser~Nam Lim.
\newblock Fedsr: A simple and effective domain generalization method for federated learning.
\newblock \emph{Advances in Neural Information Processing Systems}, 35:\penalty0 38831--38843, 2022.

\bibitem[Petroni et~al.(2019)Petroni, Rockt{\"a}schel, Lewis, Bakhtin, Wu, Miller, and Riedel]{petroni2019language}
Fabio Petroni, Tim Rockt{\"a}schel, Patrick Lewis, Anton Bakhtin, Yuxiang Wu, Alexander~H Miller, and Sebastian Riedel.
\newblock Language models as knowledge bases?
\newblock \emph{arXiv preprint arXiv:1909.01066}, 2019.

\bibitem[Poerner et~al.(2019)Poerner, Waltinger, and Sch{\"u}tze]{poerner2019bert}
Nina Poerner, Ulli Waltinger, and Hinrich Sch{\"u}tze.
\newblock E-bert: Efficient-yet-effective entity embeddings for bert.
\newblock \emph{arXiv preprint arXiv:1911.03681}, 2019.

\bibitem[Radford et~al.(2021)Radford, Kim, Hallacy, Ramesh, Goh, Agarwal, Sastry, Askell, Mishkin, Clark, et~al.]{radford2021learning}
Alec Radford, Jong~Wook Kim, Chris Hallacy, Aditya Ramesh, Gabriel Goh, Sandhini Agarwal, Girish Sastry, Amanda Askell, Pamela Mishkin, Jack Clark, et~al.
\newblock Learning transferable visual models from natural language supervision.
\newblock In \emph{International conference on machine learning}, pages 8748--8763. PMLR, 2021.

\bibitem[Rame et~al.(2022)Rame, Dancette, and Cord]{rame2022fishr}
Alexandre Rame, Corentin Dancette, and Matthieu Cord.
\newblock Fishr: Invariant gradient variances for out-of-distribution generalization.
\newblock In \emph{International Conference on Machine Learning}, pages 18347--18377, 2022.

\bibitem[Shin et~al.(2020)Shin, Razeghi, Logan~IV, Wallace, and Singh]{shin2020autoprompt}
Taylor Shin, Yasaman Razeghi, Robert~L Logan~IV, Eric Wallace, and Sameer Singh.
\newblock Autoprompt: Eliciting knowledge from language models with automatically generated prompts.
\newblock pages 4222--4235, 2020.

\bibitem[Shu et~al.(2023)Shu, Guo, Wu, Wang, Wang, and Long]{shu2023clipood}
Yang Shu, Xingzhuo Guo, Jialong Wu, Ximei Wang, Jianmin Wang, and Mingsheng Long.
\newblock Clipood: Generalizing clip to out-of-distributions.
\newblock \emph{arXiv preprint arXiv:2302.00864}, 2023.

\bibitem[Snell et~al.(2017)Snell, Swersky, and Zemel]{SnellSZ17}
Jake Snell, Kevin Swersky, and Richard~S. Zemel.
\newblock Prototypical networks for few-shot learning.
\newblock In \emph{Advances in Neural Information Processing Systems}, pages 4077--4087, 2017.

\bibitem[Tenison et~al.(2022)Tenison, Sreeramadas, Mugunthan, Oyallon, Belilovsky, and Rish]{tenison2022fedGMA}
Irene Tenison, Sai~Aravind Sreeramadas, Vaikkunth Mugunthan, Edouard Oyallon, Eugene Belilovsky, and Irina Rish.
\newblock Gradient masked averaging for federated learning.
\newblock \emph{arXiv preprint arXiv:2201.11986}, 2022.

\bibitem[Venkateswara et~al.(2017)Venkateswara, Eusebio, Chakraborty, and Panchanathan]{venkateswara2017deep}
Hemanth Venkateswara, Jose Eusebio, Shayok Chakraborty, and Sethuraman Panchanathan.
\newblock Deep hashing network for unsupervised domain adaptation.
\newblock In \emph{Proceedings of the IEEE conference on computer vision and pattern recognition}, pages 5018--5027, 2017.

\bibitem[Wang et~al.(2021)Wang, Bai, Gao, Yuan, and Li]{wang2021unsupervised}
Qi Wang, Sikai Bai, Junyu Gao, Yuan Yuan, and Xuelong Li.
\newblock Unsupervised domain adaptive learning via synthetic data for person re-identification.
\newblock \emph{arXiv preprint arXiv:2109.05542}, 2021.

\bibitem[Yang et~al.(2019)Yang, Liu, Chen, and Tong]{yang2019federated}
Qiang Yang, Yang Liu, Tianjian Chen, and Yongxin Tong.
\newblock Federated machine learning: Concept and applications.
\newblock \emph{ACM Transactions on Intelligent Systems and Technology (TIST)}, 10\penalty0 (2):\penalty0 1--19, 2019.

\bibitem[Yao et~al.(2023)Yao, Zhang, and Xu]{yao2023KCoOp}
Hantao Yao, Rui Zhang, and Changsheng Xu.
\newblock Visual-language prompt tuning with knowledge-guided context optimization.
\newblock In \emph{Proceedings of the IEEE/CVF Conference on Computer Vision and Pattern Recognition}, pages 6757--6767, 2023.

\bibitem[Yao et~al.(2022)Yao, Bai, Zhang, Zhang, Sun, Chen, Li, and Yu]{yao2022pcl}
Xufeng Yao, Yang Bai, Xinyun Zhang, Yuechen Zhang, Qi Sun, Ran Chen, Ruiyu Li, and Bei Yu.
\newblock Pcl: Proxy-based contrastive learning for domain generalization.
\newblock In \emph{Proceedings of the IEEE/CVF Conference on Computer Vision and Pattern Recognition}, pages 7097--7107, 2022.

\bibitem[Zhang et~al.(2021)Zhang, Lei, Shi, Huang, and Chen]{zhang2021fedADG}
Liling Zhang, Xinyu Lei, Yichun Shi, Hongyu Huang, and Chao Chen.
\newblock Federated learning with domain generalization.
\newblock \emph{arXiv preprint arXiv:2111.10487}, 2021.

\bibitem[Zhou et~al.(2021)Zhou, Yang, Qiao, and Xiang]{zhou2021domain}
Kaiyang Zhou, Yongxin Yang, Yu Qiao, and Tao Xiang.
\newblock Domain generalization with mixstyle.
\newblock \emph{arXiv preprint arXiv:2104.02008}, 2021.

\bibitem[Zhou et~al.(2022{\natexlab{a}})Zhou, Liu, Qiao, Xiang, and Loy]{zhou2022domain}
Kaiyang Zhou, Ziwei Liu, Yu Qiao, Tao Xiang, and Chen~Change Loy.
\newblock Domain generalization: A survey.
\newblock \emph{IEEE Transactions on Pattern Analysis and Machine Intelligence}, 2022{\natexlab{a}}.

\bibitem[Zhou et~al.(2022{\natexlab{b}})Zhou, Yang, Loy, and Liu]{zhou2022CoCoOp}
Kaiyang Zhou, Jingkang Yang, Chen~Change Loy, and Ziwei Liu.
\newblock Conditional prompt learning for vision-language models.
\newblock In \emph{Proceedings of the IEEE/CVF Conference on Computer Vision and Pattern Recognition}, pages 16816--16825, 2022{\natexlab{b}}.

\bibitem[Zhou et~al.(2022{\natexlab{c}})Zhou, Yang, Loy, and Liu]{zhou2022CoOp}
Kaiyang Zhou, Jingkang Yang, Chen~Change Loy, and Ziwei Liu.
\newblock Learning to prompt for vision-language models.
\newblock \emph{International Journal of Computer Vision}, 130\penalty0 (9):\penalty0 2337--2348, 2022{\natexlab{c}}.

\end{thebibliography}
}


\end{document}

\title{Supplementary Material for DiPrompT: Disentangled Prompts for \\ Multiple Latent Domain Generalization in Federated Learning}  % **** Enter the paper title here

\maketitle
\thispagestyle{empty}
\appendix

%%%%%%%%% BODY TEXT - ENTER YOUR RESPONSE BELOW
\section{Pseudo Code for DiPrompT}
We present the pseudo-code for DiPrompT during training time, where Q-Prompt as well as G-Prompts and D-Prompts are alternatively updated in each local client, while G-Prompts and D-Prompts are aggregated in server-side with different manners. The pseudo-code can be found in Algorithm \ref{Alg1}

\section{Dataset}
\noindent\textbf{VLCS} \cite{fang2013unbiased} is another public image classification dataset, which aggregates instances from VOC2007, LabelMe, Caltech10, and SUN09 as 4 separated domains. It contains 10, 729 instances with 5 classes. Similar to PACS and Officehome, we adopt the ``leave-one-domain-out" strategy and implement training in divided training examples from three domains and testing in the test set of the rest domain.

\noindent\textbf{PACS} \cite{li2017deeper} is a dataset containing 9,991 RGB images of 227x227 resolutions. It consists of four domains, namely Sketch (3,929 images), Cartoon (2,344 images), Art Painting (2,048 images), and Photo (1,670 images). Each domain contains seven categories: dog, elephant, giraffe, guitar, horse, house, and person. We follow the official guidance to split training and test data. In our DiPrompT training, we chose three domains for training and allocate their training data to $K$ clients, where the training data from a single domain can spread to multiple clients but a client only contains data from a single domain. Moreover, the global model tests in the test set of the rest subset.

\noindent\textbf{Officehome} \cite{venkateswara2017deep} is a larger image classification dataset, which contains around 15,500 images. It consists of images from 4 different domains: Artistic, Clip Art, Product images, and Real-World. For each domain, the sub-dataset contains images of 65 object categories found typically in office and home scenarios, such as hammer, chair, bike, table, knife and etc.  Moreover, we split the training-test subset while assigning them to sever and different clients in Officehome with a similar manner to PACS.

%-------------------------------------------------------------------------

\section{Addtional Ablation Study}
\subsection{Domain Distance.} To validate the significant differences in domain shift among various domains, we describe the distances between different domains by the average inter-class distance and cosine distance of domain centroids in the PACS dataset. As depicted in Table \ref{distance}, we observe a close relationship between Art and Photo, while the distance between Art and Sketch is substantial. Hence, when Art is designated as the target domain on the central server, local data from Photo can make more contributions than ones from Sketch. These findings support our assumption that some source domains may contain more valuable information for the target domain than other domains. Moreover, our investigation entails an examination of the correlation between domain distance and the effectiveness of domain generalization. As shown in Table \ref{performance}, we can observe a noteworthy trend: a reduction in domain distance corresponds to generalization performance improvements. The findings demonstrate that various source domains exert disparate influences on the inference performance within the target domain. Additionally, "overall" means involves harnessing datasets from other domains as training data, thereby facilitating the utilization of valuable complementary information derived from G-Prompt and D-Prompt and improving inference performance within the target domain. 

% \subsection{Detailed Results for Each domain} We provide concrete results for each domain with each dataset for ablation study in the main text, where results on few-shot recognition are shown in Table \ref{fewshot}, impacts of backbone architecture are illustrated in Table \ref{backbones} and we show results in terms of the number of clients in Table \ref{clients}.

\subsection{One client containing multi-domains data.}
% Results with multi-domain data in one client.}
%As shown in Table \ref{mul_domain}, DiPrompT performs better than other methods when one client contains data from multiple source domains, confirming its effectiveness for various federated domain generalization tasks.  %which affirms the general ability of DiPrompT to various federated domain generalization tasks.
We investigate the efficacy of DiPrompT when one client contains data from multiple source domains in Figure \ref{mul_domain}. Although all methods inevitably suffer from performance degradation due to domain shift between mixture domains within a client, DiPrompT exhibits superior performance compared to other state-of-the-art methods. These findings affirm the general ability of DiPrompT to various federated domain generalization tasks.

\subsection{The performance on seen domains.}
As shown in Table \ref{seen_domains},
we demonstrate the superior performance of DiPrompT compared to other methods, where prompts are trained and tested within the same domains.
It can be observed that all approaches exhibit enhanced performance attributed to the elimination of domain shift between the training and testing phases. Notably, our DiPrompT still outperforms other methods.

\subsection{The potential influence of class heterogeneity.}
% To avoid class heterogeneity interference, experiments were conducted under an `IID' data setting. Specifically,  we separate data into distinct categories, and evenly distribute data from each category to all clients.  Table~\ref{IID} showcases that our DiPrompt alleviates the problem of domain generalization rather than class heterogeneity.
% Furthermore, our experimental setting allows a domain to spread into multiple clients and a client with only one domain. We also provide results about a client with multiple domains in Table \ref{mul_domain}.
To avoid the interference of class heterogeneity, we conducted experiments under an ``IID" data setting. Specifically, when dividing the data of a domain into multiple clients, we initially separated data into distinct categories. Subsequently, we evenly distribute data from each category to different clients, ensuring comprehensive coverage of all categories and the quantities from different categories are close in each client. Table \ref{IID} showcases that our DiPrompt alleviates the problem of domain generalization rather than class heterogeneity.

\begin{table}[t]
	\begin{center}
		%\begin{spacing}{0.95}
		
% 		The best results are highlighted with bold fonts.
            
		\resizebox{1.0\hsize}{!}{
			\begin{tabular}{c|c|c| c| c}
				%\toprule[1.3pt]
				%\multicolumn{2}{c|}{\multirow{2}{*}
                    \hline
				\multirow{2}{*}{Domains}  & \multicolumn{4}{c}{PACS}\\
				\cline{2-5} 
				  & {Art} & {Cartoon} & {Photo} & {Sketch} \\
				\hline \hline
                    %Baseline &90.31  &92.79  &99.44 &78.35  &90.22  \\ 
                    Art   &- &355.89  &70.59  &1878.98 \\
                    Cartoon &337.52 &- &510.78 &1189.07      \\
                    Photo &28.78 &441.94 &- &2101.81    \\
                    Sketch &1761.94 &973.19 &1941.168  &- \\ \hline 	
			\end{tabular}
		}
  \caption{Distance between different domains in the PACS dataset using features from pre-trained ViT-B/16.}\label{distance}
		%\end{spacing}
	\end{center}
\end{table}
\begin{table}[t]
	\begin{center}
		
		%\begin{spacing}{0.95}
		
% 		The best results are highlighted with bold fonts.
            
		\resizebox{1.0\hsize}{!}{
			\begin{tabular}{c|c|c| c| c|c}
				%\toprule[1.3pt]
				%\multicolumn{2}{c|}{\multirow{2}{*}
                    \hline
				\multirow{2}{*}{Domains}  & \multicolumn{4}{c}{PACS}\\
				\cline{2-6} 
				  & {Art} & {Cartoon} & {Photo} & {Sketch}  & {Overall}\\
				\hline \hline
                    %Baseline &90.31  &92.79  &99.44 &78.35  &90.22  \\ 
                    Art   &- &94.68  &94.82  &93.93 &94.97 \\
                    Cartoon &95.67 &- &94.55 &95.10 &96.25     \\
                    Photo &99.52 &99.46 &- &99.26   &99.56 \\
                    Sketch &82.15 &81.97 &83.45  &- &84.72 \\ \hline 	
			\end{tabular}
		}
  \caption{Performance comparison of our proposed DiPrompT under different domains of PACS 
 dataset.}\label{performance}
		%\end{spacing}
	\end{center}
\end{table}

\begin{table}[t]
	\begin{center}
		%\begin{spacing}{0.95}
		
% 		The best results are highlighted with bold fonts.
            
		\resizebox{1.0\hsize}{!}{
			\begin{tabular}{c|c|c| c| c|c}
				%\toprule[1.3pt]
				%\multicolumn{2}{c|}{\multirow{2}{*}
                    \hline
				\multirow{2}{*}{Methods}  & \multicolumn{5}{c}{PACS}\\
				\cline{2-6} 
				  & {Art} & {Cartoon} & {Photo} & {Sketch} & {Avg}\\
				\hline \hline
                    PromptFL &91.24  &93.90 &99.22  &79.51  &90.96   \\
                    FedCLIP &91.65  &93.26 &99.28  &80.80 &91.24  \\
                    \textbf{DiPrompT} &93.46 &95.56 &99.40 &81.08 &92.37 \\ \hline 	
			\end{tabular}
		}
  \caption{Analysis for multi-domain data in one client.}\label{mul_domain}
		%\end{spacing}
	\end{center}
\end{table}

\begin{table}[t]
	\begin{center}
		%\begin{spacing}{0.95}
		
% 		The best results are highlighted with bold fonts.
		\resizebox{1.0\hsize}{!}{
			\begin{tabular}{c|c|c| c| c|c}
				%\toprule[1.3pt]
				%\multicolumn{2}{c|}{\multirow{2}{*}
                    \hline
				\multirow{2}{*}{Methods}  & \multicolumn{5}{c}{PACS}\\
				\cline{2-6} 
				  & {Art} & {Cartoon} & {Photo} & {Sketch} & {Avg}\\
				\hline \hline
                    %Baseline &90.31  &92.79  &99.44 &78.35  &90.22  \\ 
                    PromptFL &95.22  &97.23  &99.58  &89.43  &95.26   \\
                    FedCLIP &95.09  &97.14 &99.58  &89.23 &95.36  \\
                    \textbf{DiPrompT} &96.19 &97.40 &99.64 &89.74 &95.74 \\ \hline 	
			\end{tabular}
   
		}
  \caption{Quantitative analysis on seen domains.}\label{seen_domains}
		%\end{spacing}
	\end{center}
\end{table}

\begin{table}[t]
	\begin{center}
		%\begin{spacing}{0.95}
		\resizebox{1.0\hsize}{!}{
			\begin{tabular}{c|c|c| c| c|c}
				%\toprule[1.3pt]
				%\multicolumn{2}{c|}{\multirow{2}{*}
                    \hline
				\multirow{2}{*}{Methods}  & \multicolumn{5}{c}{PACS}\\
				\cline{2-6} 
				  & {Art} & {Cartoon} & {Photo} & {Sketch} & {Avg}\\
				\hline \hline
                    %Baseline &90.31  &92.79  &99.44 &78.35  &90.22  \\ 
                    PromptFL &94.31 &96.37  &99.64  &85.16  &93.87   \\
                    FedCLIP &94.04  &96.05 &99.58  &84.47 &93.53  \\
                    \textbf{DiPrompT} &95.43 &97.10 &99.70 &87.58 &94.95 \\ \hline 	
			\end{tabular}
   
		}%
  \caption{Quantitative analysis on IID data settings.}\label{IID}
		%\end{spacing}
	\end{center}
\end{table}

% \textbf{Relationship between} We also analyze the efficiency of DiPrompT in terms of the computation and communication cost during training in Figure \ref{costs}. We measure the communication cost by the size of uploaded data per round. It can be observed that DiPrompt can save at most 830 times the communication cost per round compared to FedADG and 625 times for FedGM, which makes a wide difference in communication cost between them. As for the computation cost, we report the comparison of GPU time as in the same mini-batch, where DiPrompt outperforms FedADG around 2 times and is comparable computation cost with FedGMA. It is noted that we  don't compare with other FL and FL+DG methods (i.e FedAvg, FedProx, and FedSR) since they have almost the same communication and computation costs as FedGMA, while PrompFL and FedCLIP have similar costs with our DiPrompT. These results demonstrate the efficiency of our DiPrompT, which can be applied to many real-world scenarios.

\begin{algorithm*}[t]
\algsetup{linenosize=\small} \small
	\caption{DiPrompt at training time}
	\label{alg:Framwork}
	\begin{algorithmic}[1] %这个1 表示每一行都显示数字
		\REQUIRE ~ %算法的输入参数：Input
% 		$\eta$, $\eta_s$ $\eta_w$: learning rates; 
		$K$: number of clients; $H$: number of selected clients in each round; $R$: number of training rounds; $T$: number of local iterations; $\mathcal{D}_i$: $i$-th local data;   $M$: number of source domains; $V^G$: G-Prompt, $V^D$;  D-Prompts: $V^Q$; Q-Prompt; \\
        \STATE Initialize $V^G= \left \{ v_1, ..., v_L\right\}$ $V^D = \left \{V^D_1,..., V^D_M \right\}$, $V^D_m = \left \{ v_1, ..., v_L, s_m\right\}$, $V^Q= \left \{ v_1, ..., v_L\right\}$
        
        \STATE \underline{\textbf{RunServer}}
		\FOR{each round $r$ from 0 to $R-1$}
		\STATE Randomly select $\{\mathcal{C}_k\}_{k=1}^H$ from $K$ clients;
            
		\FOR{each $k \in [1, H]$ \textbf{in parallel}}
            %\STATE m = M[$k$];
		\STATE $V^{G,{r+1}}$, $V^{D,{r+1}}$ $\leftarrow$ \underline{\textbf{RunClient}}($V^{G,{r}}$, $V^{D,{r}}$)
		\ENDFOR
		\STATE $V^{G,{r+1}} \leftarrow \frac{1}{\sum_{i=1}^{H} (| \mathcal{D}_i|)}\sum_{i=1}^{H} (|\mathcal{D}_i|) \cdot V_i^{G,{r+1}}$  \hfill % 
  $\vartriangleright$ G-Prompt aggregation
            \FOR{each $m$ from 0 to $M$}   
            \STATE Update from $V_m^{r}$ to $V_m^{r+1}$ with Eqn. \textcolor{red}{4}
            \STATE Update $\hat{V}_m^{r+1}$ with Eqn. \textcolor{red}{5}
		\ENDFOR \hfill $\vartriangleright$ D-Prompts aggregation
	      \ENDFOR
            
		\RETURN $V^{G,R}$, $V^{D,R}$;
		%\FOR{each $i \in [1,9]$}
        \STATE \underline{\textbf{RunClient}}($V^{G,{r}}$, $V^{D,{r}}$)
        \STATE $V_{k}^{G,(0)}$ $\leftarrow$ $V^{G,{r}}$; $V^{D,(0)}$ $\leftarrow$ $V^{D,{r}}$;
            
		\FOR{each local iteration $t$ from 0 to $T-1$}
		%\FOR{minibatch $\widetilde{\mathcal{D}}^i \in \mathcal{D}_i$}
            \STATE $\mathcal{D}^k = \left \{x_i,y_i \right \}_{k=1}^{|{\mathcal{D}}_i|}$
		\STATE Update from $V_Q^{(0)}$ to $V_Q^{(1)}$  with Eqn. \textcolor{red}{7}   %\hfill 
            \STATE Select $m$-th D-Prompt $V^D_{k,m}$ from $V^D_k$ with Eqn. \textcolor{red}{6}
  %$\vartriangleright$ step 5
            \STATE Update from $V_{k}^{G,(0)}$ to $V_{k}^{G,(1)}$ with Eqn. \textcolor{red}{2}
            \STATE Update from $V_{k,m}^{D,(0)}$ to $V_{k,m}^{D,(1)}$ with Eqn. \textcolor{red}{3}    \hfill
            $\vartriangleright$ Simultaneous update G-Prompt and D-Prompts
		%\ENDFOR
		\ENDFOR
            \STATE Send $V_{k}^{G,(T)}$ $\leftarrow$ $V^{G,{r}}$; $V^{D,(T)}$ to the server and set $V_Q^{(0)} =V_Q^{(T)}$ in local client
	\end{algorithmic}\label{Alg1}
\end{algorithm*}

\newpage

\newpage
{
    \small
    \bibliographystyle{ieeenat_fullname}
    \bibliography{main}
}